\documentclass[aoas, preprint, authoryear]{imsart}

\RequirePackage[OT1]{fontenc}
\RequirePackage{amsthm,amsmath,amssymb}
\RequirePackage[numbers]{natbib}
\RequirePackage[colorlinks,citecolor=blue,urlcolor=blue]{hyperref}
\usepackage{graphicx}
\usepackage{subcaption}

\usepackage{geometry}
\graphicspath{{./figures/}}

\newcommand{\Twiddle}{\mbox{\(\tt\sim\)}}
\newcommand{\minimize}{\operatornamewithlimits{minimize}}

\newcommand{\argmin}{\operatornamewithlimits{arg\,min}}

\DeclareMathOperator{\diag}{diag}
\DeclareMathOperator{\sign}{sgn}

\newcommand{\fhat}{\widehat{f}}
\newcommand{\R}{\mathbb{R}}
\numberwithin{equation}{section}
\theoremstyle{plain}

\def\trdinv{}
\def\subgrad{s}

\begin{document}

\begin{frontmatter}
\title{Generalized Additive Model Selection}
\runtitle{GAMSEL}
\thankstext{T1}{
Trevor Hastie was partially supported by grant DMS-1407548 from the National
Science Foundation, and grant RO1-EB001988-15 from the National Institutes of
Health.
}

\begin{aug}
\author{\fnms{Alexandra} \snm{Chouldechova}\ead[label=e1]{achould@stanford.edu}}
\and
\author{\fnms{Trevor} \snm{Hastie}\thanksref{T1}\ead[label=e2]{hastie@stanford.edu}}

\runauthor{A. Chouldechova and T. Hastie}

\affiliation{Carnegie Mellon and Stanford University}

\address{Sequoia Hall\\
390 Serra Mall\\
Stanford, CA 94305\\ 
\printead{e1}\\
\phantom{E-mail:\ }\printead*{e2}}
\end{aug}

\begin{abstract}
  We introduce GAMSEL (Generalized Additive Model Selection), a
  penalized likelihood approach for fitting sparse generalized
  additive models in high dimension.  Our method interpolates between
  null, linear and additive models by allowing the effect of each
  variable to be estimated as being either zero, linear, or a
  low-complexity curve, as determined by the data.  We present a
  blockwise coordinate descent procedure for efficiently optimizing
  the penalized likelihood objective over a dense grid of the tuning
  parameter, producing a regularization path of additive models.  We
  demonstrate the performance of our method on both real and simulated
  data examples, and compare it with existing techniques for additive
  model selection.
\end{abstract}



\end{frontmatter}

\section{Introduction}

In many applications it may be too restrictive to suppose that the
effect of all of the predictors is captured by a simple linear fit of
the form, $\eta(x_i)=\sum_{j=1}^p \beta_j x_{ij}$.
Generalized additive models, introduced in
\cite{hastie1986generalized}, allow for greater flexibility by
modeling the linear predictor of a generalized linear model as a sum
of more general functions of each variable:
\[
 \eta(x_i) = \sum_{j=1}^p f_j(x_{ij}),
\]
where the $f_j$ are unknown functions, assumed to be smooth or otherwise low-complexity.  While generalized additive models have become widely used since their introduction, their applicability has until recently been limited to problem settings where the number of predictors, $p$, is modest relative to the number of observations, $n$.  In this paper we propose a method for generalized additive model selection and estimation that scales well to problems with many more predictors than can be reasonably handled with standard methods.

In large data settings it is often fair to assume that a large number
of the measured variables are irrelevant or redundant for the purpose
of predicting the response.  It is therefore desirable to produce
estimators that are sparse, in the sense that $\hat f_j \equiv 0$ for
some, or even most, predictors.  Furthermore, in many applications it
may be reasonable to assume that the linear model, $f_j(x) = \beta_j
x$, is adequate for many of the predictors.  Linear relationships are
easy to interpret, and the widespread use of linear models suggests
that linear fits are often believed to be good approximations in
practice. For purely linear models, the lasso \citep{tibshirani1996regression} is an effective form of regularization that performs model selection. The package {\tt glmnet} \citep{friedman08:_regul_paths_gener_linear_model_coord_descen} implements the lasso regularization path for an important subset of the class of generalized linear models.

Our proposed estimator selects between fitting each $f_j$ as zero, linear, or nonlinear, as determined by the data.  In so doing it retains the interpretability advantages of linear fits when appropriate, while capturing strong non-linear relationships when they are present.  Our method, which we call GAMSEL (Generalized Additive Model Selection), is based on optimizing a penalized (negative) log-likelihood criterion of the form,
\begin{equation*}
 \hat f_1, \ldots, \hat f_p = \argmin_{f_1,\ldots, f_p\in \mathcal{F}} \,\, \ell(y;f_1, \ldots, f_p) + \sum_{j = 1}^p J(f_j)
\end{equation*}
for a particular choice of $J(f)$.  We give the form of the penalty term in Section~\ref{sec:objective}, after presenting some motivating preliminaries.  


\subsection{Related literature}

Since the introduction of the lasso in \cite{tibshirani1996regression}, a significant body of literature has emerged establishing the empirical and theoretical success of $\ell_1$ penalized regression in high-dimensional settings.  The method we introduce in this paper can be viewed as one possible extension of the lasso to the additive model setting.  

There have been numerous other attempts at this extension.  The methods most closely related to our proposal are COSSO \citep{lin2006component}, SpAM \citep{ravikumar2007spam}, and the method of \citet{meier2009high}.  Each of these proposals is based on penalized likelihood and the difference comes from the choice of $\mathcal{F}$ and the structure of the penalty term~$J(f)$.  COSSO models the components $f_j$ as belonging to a reproducing kernel Hilbert space (RKHS), and operates by penalizing the sum of the component RKHS-norms (instead of the sum of squared norms).  SpAM is effectively a functional version of the group lasso \citep{yuan2006model}; it decouples the choice of smoother from the sparsity constraint, and is thus broadly applicable to any choice of smoothing operator.  The penalty function proposed in \cite{meier2009high} is the quadratic mean of the component function norm and a second derivative smoothness penalty, summed over the components.  The authors argue that the quadratic mean penalty structure enjoys both theoretical and computational advantages that closely related formulations do not.

Our method is distinct from the existing proposals in that it selects between linear and non-linear fits for the component functions.  Later in the paper we present experimental results showing that GAMSEL performs nearly as well as SpAM when the true component functions are all non-linear, and can perform considerably better when some of the true component functions are linear.  

During the writing of this manuscript we became aware of the SPLAM method being developed independently by Lou, Bien, Caruana, and Gehrke \citep{lou2014sparse}.  This method also selects between zero, linear and non-linear fits for component functions in a generalized additive model framework, but differs in the choice of penalty function.  

Model selection is of course a very old problem in statistics, and there are several popular methods for variable and smoothness selection that were developed without the high-dimensional or sparse setting in mind.  Of particular note is the AIC-based stepwise selection procedure implemented in the {\tt gam} R package \citep{gampkg}, as well as the methods implemented in the widely used {\tt mgcv} package \citep{wood2011fast}.  Later in the paper we compare the selection performance of GAMSEL and {\tt step.gam}.  The methods implemented in {\tt mgcv} have a different focus in that they are geared more for smoothness selection rather than variable selection.

\subsection{Layout of the paper}
We begin in Section 2 by giving a brief overview of smoothing splines
in order to motivate the form of the penalty term in the GAMSEL
objective.  The full form of the penalized likelihood objective is
introduced in Section~\ref{sec:objective}.  Sections~\ref{sec:generating-bases} and \ref{sec:blockw-coord-desc} present
technical details concerning the fitting procedure.  Section~\ref{sec:generating-bases} 
describes the basis and penalty matrix construction used by the
procedure, and Section~\ref{sec:blockw-coord-desc} presents the blockwise coordinate descent
algorithm for fitting the GAMSEL model.  In Section~\ref{sec:experiments} we investigate
the model selection performance of our method on a simple simulated
example, and present the results of fitting GAMSEL to the Boston
Housing and HP spam datasets.  In Section~\ref{sec:comp-other-meth} we compare the model
selection and prediction performance of GAMSEL to that of two
competing methods: SpAM and {\tt step.gam}. Section~\ref{sec:conclusion} wraps the paper up with some conclusions.

%
%

\section{Method}  In this section we describe the penalized optimization problem underlying GAMSEL. We begin with a brief review of smoothing splines to help motivate the form of the penalty term.  This is followed by a brief overview of generalized additive models. For a more comprehensive treatment of both of these subjects we refer the reader to \cite{green1993nonparametric}, \cite{hastie1990generalized} or \cite{wood2011fast}.

\subsection{Overview of Smoothing Splines}  Consider the univariate regression setting where we observe data on a response variable $y$ and a single predictor $x$.  We assume that $x$ takes values in some compact interval on the real line.  Given observations $(y_1, x_1), \ldots, (y_n, x_n)$ and a choice of $\lambda \ge 0$, the smoothing spline estimator, $\fhat_\lambda$, is defined by
\begin{equation} 
  \fhat_\lambda = \argmin_{f \in C^2} \sum_{i = 1}^n (y_i - f(x_i))^2 + \lambda \int f''(t)^2 dt 
  \label{eq:smoothspline}
\end{equation}
While at first glance this may appear to be a difficult optimization problem due to the uncountably infinite parameter space $C^2$, it can be shown that problem reduces to a much simpler form.  The solution to \eqref{eq:smoothspline} is a natural cubic spline with knots at the unique values of $x$.  It can be shown that the space of such splines is $n$-dimensional, which allows us to express the smoothing spline objective as a finite dimensional problem.   

Given a set of natural spline basis functions $\{h_1, h_2, \ldots, h_n\}$, we can re-express the objective function $f$ $f$ as $f(x) = \sum_{j=1}^nh_j(x)\theta_j$.  Forming $H_{n\times n} = \{h_j(x_i)\}_{i,j=1}^n$ we note that solving \eqref{eq:smoothspline} amounts to estimating $\theta$ according to,
\begin{equation}
  \hat \theta_\lambda = \argmin_{\theta \in \R^n} \| y - H \theta \|_2^2 + \lambda \theta^T \Omega \theta
  \label{eq:ridgeform}
\end{equation}
where the \emph{penalty matrix} $\Omega$ has entries $\Omega_{jk} = \int h_j''(x)h_k''(x)dx$.  Note that $\Omega$ depends on the $h_j$, which in turn depend on the $x_i$, but not on $y$ or $\theta$.  This is just a generalized ridge regression problem, so we know that the the $n$-vector of fitted values\footnote{We are overloading notation here by using $\hat f_\lambda$ to denote both the full smoothing spline function and the $n\times 1$ vector of fitted values.  } has the form,
\[
 \hat f_\lambda = H\hat \theta_\lambda = S_\lambda y.
\] 
The matrix $S_\lambda$ appearing above is commonly referred to as the \emph{smoother matrix}.  To better understand the action of $S_\lambda$, it helps to take a special choice of spline basis.   

While the penalty matrix $\Omega$ is in general not diagonal, it
becomes diagonal if we take our basis to be the \emph{Demmler-Reinsch
  basis} \citep{demmler75:_oscil_matric_with_splin_smoot}, which we'll denote by $\{u_1(x), u_2(x), \ldots, u_n(x)\}$.
We'll also denote the now-diagonal penalty matrix by $D = \diag(d_1,
d_2, \ldots, d_n)$.  Assuming that the $u_j$ are ordered in increasing
order of complexity, $D$ has the property that $0 = d_1 = d_2 < d_3
\le \ldots \le d_n$.  $d_1$ and $d_2$ correspond to the constant and
linear basis functions, respectively.  The non-zero $d_j$ are
associated with $u_j$ that are non-linear functions of $x$, with
higher indexes corresponding to $u_j$ with greater numbers of
zero-crossings. Furthermore, the $n\times n$ matrix $U$ with columns the $u_j$ is orthonormal.\footnote{We can find a matrix $A$ such $U=HA$ is orthonormal, and $A^T\Omega A$ is diagonal, and elements increasing.}

In the new choice of basis, the estimation problem~\eqref{eq:ridgeform} reduces to,
\begin{align}
    \hat \theta_\lambda &= \argmin_{\theta \in \R^n} \| y - U \theta \|_2^2 + \lambda \theta^T D \theta \label{eq:reinsch} \\
    &= (I + \lambda D)^{-1}U^T y \notag
\end{align}
With further rewriting, we have that the smoothing spline solution $\fhat_\lambda$ takes the form,  
\begin{align}
  \fhat_\lambda &= \underbrace{U(I + \lambda D)^{-1} U^T}_{S_\lambda}y 
  = \sum_{j = 1}^n u_j \frac{\langle u_j, y \rangle}{1 + \lambda d_j}
  \label{eq:mainspline}
\end{align}
Since $d_1 = d_2 = 0$, we see from \eqref{eq:mainspline} that $S_\lambda$ has the effect of keeping the components along the first two basis functions intact (i.e., unpenalized) and shrinking the later components by a factor of $\tau_j = \frac{1}{1 + \lambda d_j} < 1$. The \emph{degrees of freedom} for a smoothing spline is defined to be $\mbox{df}_\lambda = \sum_{j=1}^n\tau_j=\mbox{tr}(S_\lambda)$, and lies between $2$ and $n$.

Using this insight we can rewrite \eqref{eq:reinsch} by separating out the constant term $u_1$ and linear term $u_2$ to obtain the optimization problem
\begin{equation}
  \label{eq:unipls}
    (\hat \alpha_0, \hat \alpha, \hat \beta) = \argmin_{\alpha_0 \in \R, \alpha \in \R, \beta \in \R^{n-2}}\| y - \alpha_0  - \alpha x - U_{3:n}\beta\|_2^2 + \lambda \beta^T D_{3:n} \beta,
\end{equation}
where $U_{3:n}$ is the matrix with columns $u_3, \ldots, u_n$ and $D_{3:n} = \diag(d_3, d_4, \ldots, d_n)$.  
So the smoothing spline fits a mix of terms linear in $x$ and nonlinear in $x$, and the nonlinear terms are penalized by an amount that increases with their complexity.
It is this formulation of the problem that most directly motivates the GAMSEL objective we introduce in Section~\ref{sec:objective}
%


%

\begin{figure}
\centering
\includegraphics[width=\textwidth]{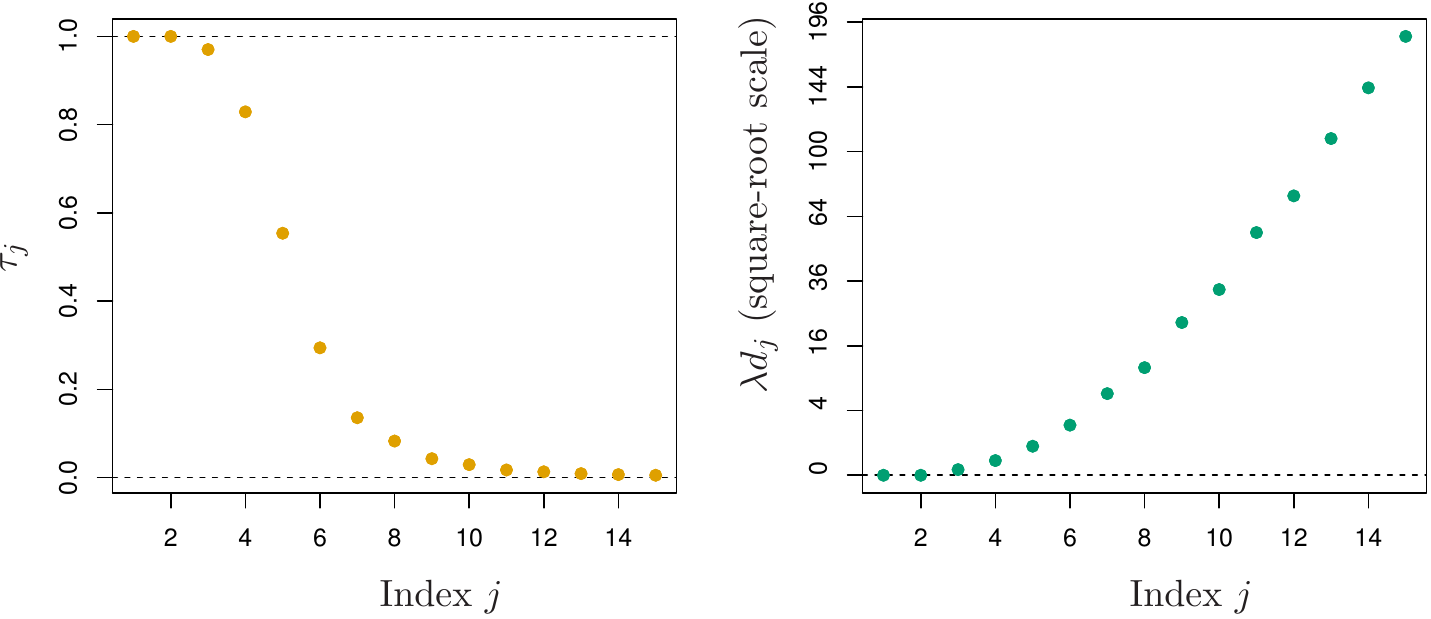}
\caption{Values of the first $15$ shrinkage factors $\tau_j$ (left) and adjusted penalty values $\lambda d_j$ (right) for a smoothing spline with $5$ degrees of freedom.  }
\label{fig:penvals}  
\end{figure}

Before moving on, it is instructive to look at the rate of decay of the shrinkage factors $\tau_j$, which are displayed along with the corresponding $\lambda d_j$ in Figure~\ref{fig:penvals} for a smoothing spline with $5$ degrees of freedom.  As we can see, the shrinkage parameters tend to decay rapidly to $0$, which means that high order basis functions generally have very little effect on the model fit.  This suggests that we would not lose much by replacing (\ref{eq:reinsch}) with a truncated version where we only consider the first $m$ basis functions for $m \ll n$.  For instance, in the case of 5 degrees of freedom we may be comfortable proceeding with $m=10$ basis functions.


\subsection{Overview of Generalized Additive Models}
\label{sec:gams}
Although generalized additive models \citep{hastie1990generalized} are defined for general smoothers and distribution families, we will limit our discussion to the use of smoothing splines and the Gaussian model. Consider the extension of criterion~(\ref{eq:smoothspline}) when there are $p$ predictors
\begin{equation}
  \label{eq:amspl}
  \minimize_{\{f_j\in C^2\}_1^p}\sum_{i=1}^n(y_i-\sum_{j=1}^pf_j(x_{ij}))^2 +\lambda \sum_{j=1}^p\int f_j''(t)^2dt. 
\end{equation}
Using extensions of the arguments for smoothing splines, one can show that the solution is finite dimensional, and each function $\hat f_j$ can be represented in a natural spline basis in its variable $x_j$. Since each function includes an intercept, we typically isolate the intercept, and insist that each fitted function average zero over the training data (without any loss of generality).
With some simplification of the notation in~(\ref{eq:unipls}), this leads to the optimization problem
 \begin{equation}
  \label{eq:multipls}
    \minimize_{\alpha_0 \in \R,\; \{\beta_j \in \R^{n-1}\}_1^p}\| y - \alpha_0  - \sum_{j=1}^pU_j\beta_j\|_2^2 + \lambda \sum_{j=1}^p\beta_j^T D_j \beta_j.
\end{equation}
The matrix $U_j$ represents an orthonormal Demmler-Reinsch basis for
variable $x_j$; each of its columns are mean centered (orthogonal to
the column of ones), and the first column is a unit-norm version of
$x_j$. $D_j$ is a diagonal penalty matrix as in the previous section,
except only the first element is zero (since the intercept term is not
represented here).

In principle this problem is a large generalized ridge regression,
although naive computation would lead to an $O((np)^3)$
algorithm. Since a single smoothing spline can be implemented in
$O(n)$ computations, \cite{hastie1990generalized} proposed the
backfitting algorithm for fitting the model. This is a block coordinate
descent approach, with $p$ blocks, each costing $O(n)$
computations. Hence the entire algorithm is $O(knp)$, where $k$ is the
number of backfitting loops required for convergence.
\cite{wood2011fast} instead reduces the dimensionality of each basis $U_j$ as eluded to in the previous section, and hence reduces the computations of the generalized ridge regression.
\cite{marx98:_direc_gener_addit_model_with_penal_likel} also use a truncated bases of \emph{P-splines} to represent penalized splines and generalized additive models.

Both the {\tt gam} package \citep{gampkg} and the {\tt mgcv} package
\cite{mgcvpkg} in {\tt R} can fit GAMs with smoothing splines. The
\texttt{step.gam} function in \texttt{gam} allows the use to perform
model selection via a step-wise procedure. For each variable, the user
provides an ordered list of possibilities, such as \emph{zero, linear,
  smooth with 5df, smooth with 8df}. Starting from say the null model
(where all terms are zero), and in \emph{forward} mode, the procedure
tries a move up the list for each term, and picks the best move (in
terms of AIC). While quite effective, this tends to be laborious and
does not scale well. The function \texttt{gam.selection} in package
\texttt{mgcv} is designed to select separately the correct amount of
smoothing for each term, rather than feature selection.


\subsection{GAMSEL Objective Function} \label{sec:objective}  With this motivation in mind, we now present our criterion for fitting a GAM with built-in selection.  For ease of presentation we focus here on the case of squared-error loss.  The extension to the logistic regression setting is presented later in Section~\ref{sec:logistic}.

We have data $(y_i, x_i)$ for $i=1, \ldots, n$.  We represent the mean function for the $j$th variable as a low-complexity curve of the form
\[ f_j(x) = \alpha_j x_j + u_j(x_j)^T\beta_j, \]
where $u_j$ is a vector of $m_j$ basis functions.
Let $U_j \in \R^{n\times m_j}$ be the matrix of evaluations of this function at the $n$ values $\{x_{ij}\}_1^n$, and assume without loss of generality that $U_j$ has orthonormal columns. 

GAMSEL estimates the $f_j$ by solving the convex optimization problem
\begin{multline}
  \label{eq:objective}
\text{minimize}_{\alpha_0, \{\alpha_j\}, \{\beta_j\}} \frac{1}{2} \left\| y - \alpha_0 - \sum_{j = 1}^p \alpha_j x_j - \sum_{j = 1}^p U_j \beta_j \right\|_2^2\\
 + \lambda\underbrace{ \sum_{j = 1}^p\left(\gamma |\alpha_j| 
+ (1-\gamma)  \|\beta_j\|_{D_j^*} \right)}_{\text{selection penalty}}\quad
+ \quad \frac{1}{2}\underbrace{\sum_{j = 1}^p \psi_j \beta_{j}^T D_j \beta_{j} }_{\text{end-of-path penalty}}
\end{multline}
where $ \|\beta_j\|_{D_j^*}  = \sqrt{\beta_j^TD_j^*\beta_j}$.
We have taken some liberties with the notation in (\ref{eq:objective}); both  $y$ and $x_j$ are now $n$ vectors. 

Let us first focus on the \emph{end-of-path penalty}, which is all that is enforced when  $\lambda=0$.
The criterion is now equivalent to (\ref{eq:amspl}) (even though the linear terms are represented twice!).
The multiplier $\psi_j$ for each term is chosen so that the fit for that term alone  would result in a  pre-specified degrees of freedom. Hence when $\lambda=0$, we fit a generalized additive model with pre-specified degrees of freedom for each term.

The \emph{selection penalty} is more complex, and consists of a
mixture of 1-norm and 2-norm penalties for each term. These take the
form of an overlap group-lasso penalty \citep{jacob2009group}, which
has the effect of inducing sparsity in the fitted model. 
The term  $\|\beta_j\|_{D_j^*}$ is a group-lasso penalty \cite{yuan2006model}; it behaves like the lasso but for a vector of coefficients. This penalty either includes all the parameters, or sets them all to zero. The overlap refers to the fact that each $x_j$ has a a pair of linear coefficients, one represented in  $ \|\beta_j\|_{D_j^*}$, the other in $|\alpha_j|$.
Here the
matrix $D_j^*$ is identical to $D_j$, except the $0$ in position one
is replaced by a $1$ (so the linear term in $U_j$ is penalized); we
discuss these $D_j^*$ in more detail later in this section. The
parameter $\gamma$ is between $0$ and $1$, and allows one to favor
linear terms (small $\gamma$) over non-linear terms, or vice
versa. Due to the particular structure of this penalty, there are
three possibilities for each predictor.

\begin{figure}[tp]
  \centering
  \includegraphics[width=\textwidth]{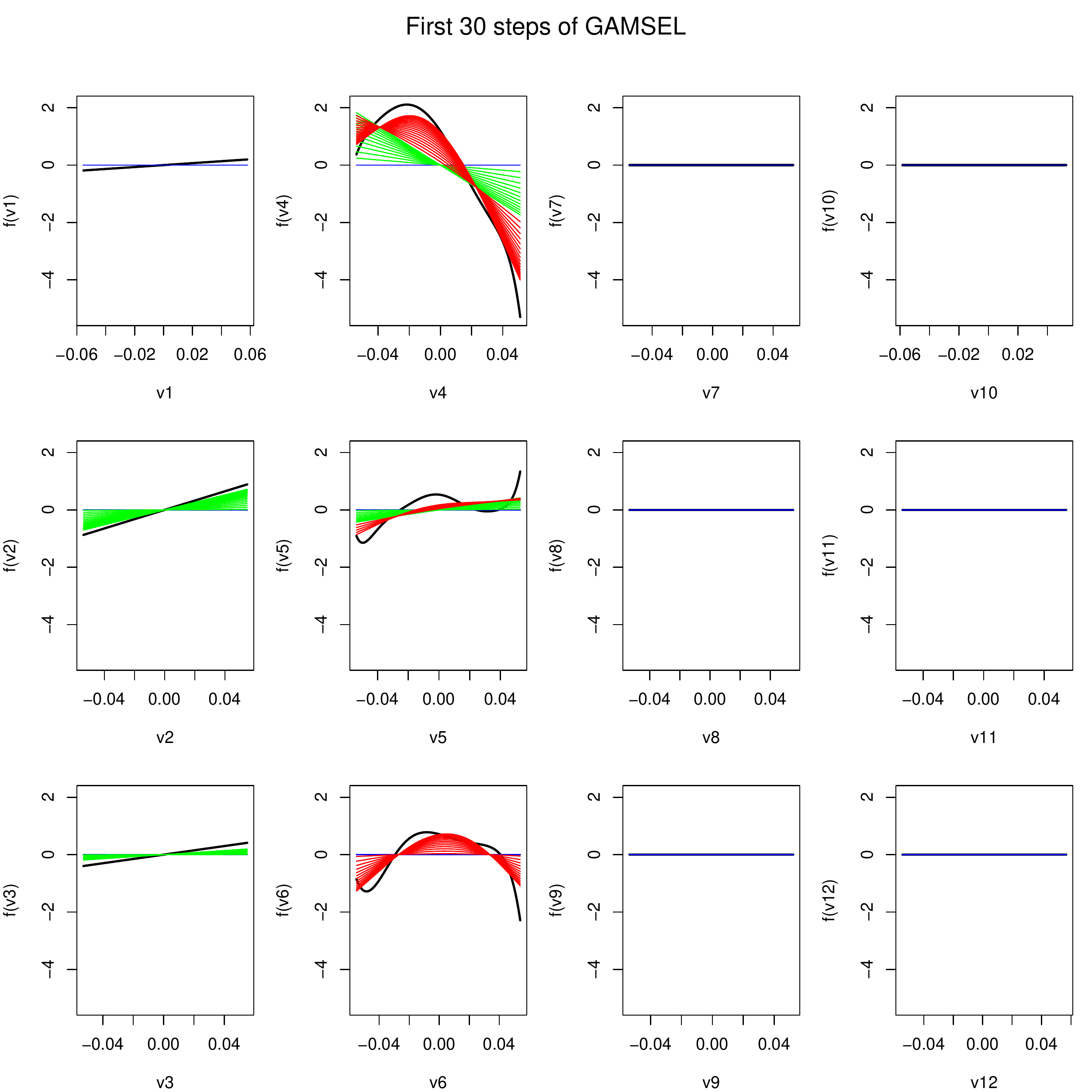}
  \caption{Synthetic example with 12 variables, one panel per variable, to illustrate the \emph{sticky} nature of the GAMSEL procedure. Gaussian noise was added to an additive model in these twelve uniformly-distributed variables. The true functions are shown in black. In the first column, the three true functions are  linear, in the second nonlinear, and in the remaining two columns all six terms are zero. We show the progression (via overplotting) of the GAMSEL procedure in its first 30 steps (out of 50), corresponding to decreasing values of $\lambda$. The estimated functions are in color:  blue means zero, green means linear, and red nonlinear. We see that 7 functions are exactly zero, two linear terms are well approximated, and the nonlinear terms are reasonably well approximated, some better than others. }
  \label{fig:sticky}
\end{figure}

\begin{description}
  \item[{\it Zero} $(\alpha_j =0, \beta_j \equiv 0)$. ] For large values of the penalty parameter $\lambda$, the penalty term can dominate the lack-of-fit term, which results in the minimizer having both $\alpha_j =0$ and $\beta_j\equiv0$.  This corresponds to the case $f_j(x) \equiv 0$.
  \item[{\it Linear} $(\alpha_j \neq 0, \beta_j \equiv 0)$. ]  For moderate values of the parameter $\lambda$ and sufficiently small $\gamma > 0$, the minimizer can have $\alpha_j \neq 0$ and $\beta_j \equiv 0$.  This corresponds to the case where $f_j(x) = \alpha_j x$ is estimated to be a strictly linear function of $x$.  
  \item[{\it Non-linear} $(\beta_j \neq 0)$. ]  For small values of $\lambda$ and/or large values of $\gamma$, the minimizer can have $\beta_j \neq 0$.  This corresponds to fitting a low-complexity curve of the form $f_j(x) = \alpha_jx + U_j \beta_j$ for the $j$th predictor.  Note that depending on the choice of $\gamma$, $\alpha_j$ may or may not be $0$, but because of the overlap (the first column of $U_j$ is a linear term), this implies a linear term is present.
\end{description}
We refer to these choices as \emph{sticky points} since certain thresholds have to be met to cross from one state to the next. Typically, as we relax $\lambda$, there is a transition from zero through linear to nonlinear, settling finally on the end-of-path nonlinear term with pre-specified degrees of freedom. Figure~\ref{fig:sticky} illustrates on a synthetic example.

Now we explain the $D_j^*$ in more detail. The $D_j$ and hence  $D_j^*$ are customized versions of the diagonal $D$ matrix from the previous section, separately for each variable $x_j$. The first element of $D_j$ is zero, and all subsequent elements are positive and increasing (see the right plot in Figure~\ref{fig:penvals}, and recall there is an extra zero there for the intercept). We have scaled these elements of $D_j$ so that the second element is $1$ (the penalty for the first nonlinear term). Hence in $D_j^*$, when we replace the zero in the first entry by a $1$, this means that as far as this penalty is concerned, a linear term is treated the same as the first nonlinear component.

With these details in mind, we can now understand the selection penalty better. If a function $f_j$ is really linear, it can enter via selection through $\alpha_j$ or the first coefficient of $\beta_j$; with $\gamma=0.5$, selection through $\beta_j$ picks up unwanted penalty on the other components of $\beta_j$ which will model the noise, and so the linear term is more likely to enter via $\alpha_j$. With $\gamma<0.5$ we can encourage this even more strongly.

In principal, once can fine-tune these penalties even further. For example, with the group lasso and our parametrization, one can argue that each penalty $\|\beta_j\|_{D_j^*}$ should be multiplied by the scalar $\phi_j=\sqrt{\mbox{tr}({D_j^*}^{-1})}$, to put each of these terms on an equal footing. Our standardization does something similar, while allowing for easier understanding of the treatment of the linear term.

We provide an R implementation for fitting this model along a dense path of values of $\lambda$ --- essentially the entire regularization path for (\ref{eq:objective}). We describe the package \texttt{gamsel} in Section~\ref{sec:gamsel.pkg}.

\section{Generating the spline bases}\label{sec:generating-bases}
In this section we discuss how we produce the basis matrix $U$ and
associated penalty matrix $D$ for any variable $x$. The idea is to approximate the truncated  Demmler-Reinsch basis and penalty for a smoothing spline. While this can be
achieved in a variety of ways, we adopt the {\em pseudo-spline}
approach of \citet{Hastie95}, which is particularly simple. It hinges on the fact that the Demmler-Reinsch basis shares the zero-crossing behaviour of orthogonal polynomials.  Let $S_\lambda$ denote the $n\times n$
smoothing-spline operator matrix for smoothing against $x$, and assume
that $\lambda$ has been chosen to achieve a fixed
$df=\mbox{trace}(S_\lambda)$. Our goal is to produce a low-rank
approximation to $S_\lambda$; that is, we want to find a $n\times k$
basis matrix $U$, and a diagonal penalty matrix $D$ such that
$S_\lambda \approx U(I+\lambda D)^{-1}U^T$.  A typical value for $k$ is 10.

Let $P$ be the $n\times k$ matrix of orthogonal polynomials generated from $x$. 
The version of pseudo-splines we use here solves
\begin{equation}
  \label{eq:ps1}
  \min_M||S_\lambda -PMP^T||_F.
\end{equation}
The solution is easily seen to be $M=P^TS_\lambda P$. We then compute the eigen-decomposition $M=VD_SV^T$, and let
$U=PV$ and $D=D_S^{-1}-I$. 

The approximation is easy to compute, since $S_\lambda P$ simply
executes the smoothing spline $k-2$ times (the first two columns of
$P$ are constant and linear, and so pass through $S_\lambda$
untouched. The smoothing spline-algorithm we employ can apply the smoother to a matrix, so the setup is not repeated for each column of $P$. See the reference for more detail.  The first two penalty
elements are $D_{11}=D_{22}=0$; the remaining elements are positive
and increasing. In practice, we discard the first element (corresponding to the intercept), and this leads to the bases $D_j$ and $U_j$ referred to in (\ref{eq:amspl}) and (\ref{eq:objective}).

The basis $U$ found by solving (\ref{eq:ps1}) lies in the space of polynomials; \citet{Hastie95} describes a simple improvement 
that requires a bit more computation. In the first step we replace $P$ by $Q$ where $QR=S_\lambda P$ (QR decomposition). Now $Q$ is an orthogonal basis in the space of natural splines in $x$. Solving (\ref{eq:ps1}) with $Q$ rather than $P$ leads to a better approximation to $S_\lambda$ (in Frobenius norm). In practice, the improvements are very small, and generally not worth the effort. Our package {\tt gamsel} allows either of these options.

\subsection{Prediction at new points}
\label{sec:predictbasis}
We have generated these bases (and penalties) for the $n$ training observations. How do we use them to make predictions at new points? It suffices to consider the case of a single variable $x$.

Armed with a basis $U$ and associated diagonal penalty matrix $D$, we fit a smooth term to a response vector $y$ by solving 
\begin{equation}
  \label{eq:ps2}
  \min_\beta ||y-U\beta||_2^2 +\lambda\beta^TD\beta,
\end{equation}
or a similar quantity in the case of logistic regression. To make predictions at a vector of new points $x_0$, we need to generate the corresponding matrix $U_0$, and then the predictions are $\hat f_0=U_0\hat\beta$. Well, $U_0=P_0V$, and $P_0$ is the matrix of orthogonal polynomials evaluated at $x_0$. If we used $Q$ rather than $P$ to generate the basis, we have to use the appropriate matrix $Q_0$ obtained by making predictions from the smoothing spline fits that produced $Q$.

\subsection{Subsetting a basis}
\label{sec:subset}
We compute all the bases for each variable $x_j$ in advance, and these are used throughout the GAMSEL computations.
Our software assumes that the $U_j$ for each variable are orthonormal matrices. When we run cross-validation to select the tuning parameter(s), it might seem that we have to regenerate the bases each time, since the subsetted $U_j$ will no longer be orthonormal.
It turns out that this is not necessary, nor desirable. We would like to work with the same basis as before, and simply modify the problem accordingly.  Let $U_1$ be the non-orthonormal matrix obtained by removing a subset $\cal F$ of of the rows of $U$, and let $y_1$ be the corresponding subset of $y$.
The full-data problem solves (\ref{eq:ps2}), which can be written as 
\begin{equation}
  \label{eq:ps2a}
  \min_\beta \frac1N||y-U\beta||_2^2 +\lambda_*\beta^TD\beta,
\end{equation}
where $\lambda_*=\lambda/N$.
It makes sense for the reduced-data estimate to solve
\begin{equation}
  \label{eq:ps3}
 \min_\beta \frac1{N_1}||y_1-U_1\beta||_2^2 +\lambda_*\beta^TD\beta.  
\end{equation}
Now this is not in the required orthogonal form, but can easily be transformed.
It is easy to show that the following steps produce a $U_1^*$ and a $D^*$ that produce exactly the solution to (\ref{eq:ps3}), but with $U_1^*$ in the required orthogonal form.
\begin{enumerate}
\item Compute the SVD $U_1D^{-\frac12}= U_1^*D_2V_2^T$.
\item $D^*=D_2^{-2}$.
\end{enumerate}
Furthermore, since $U_1^*$ is a linear transformation of $U_1$, which is itself a linear transformation of $P_1$ (corresponding subsetted version of $P$), the information for predicting from this reduced basis can be saved. Note that if the problems are solved in the unstandardized form (\ref{eq:ps2}), then the penalty $\lambda$ should be transformed to $\frac{N_1}{N}\lambda$ for the reduced problem. Note also that while this works fine for a problem with quadratic penalty, our criterion (\ref{eq:objective}) treats the linear term in a special way. However, since the linear penalty in $D$ is zero, this term is isolated and is not rotated in the construction of $U_1^*$.

\section{Blockwise Coordinate Descent Algorithm} \label{sec:blockw-coord-desc}

In this section we describe a blockwise coordinate descent algorithm for fitting the GAMSEL model in (\ref{eq:objective}).  Our routine returns parameter estimates along a sequence of lambda values, $\lambda_{\mathrm{max}} = \lambda_1 > \lambda_2 > \ldots > \lambda_L$.  We take advantage of warm starts by initializing the estimates for $\lambda_{k+1}$ at the solution for $\lambda_k$.  We also use sequential strong rules \citep{tibshirani2012strong} to provide considerable speedups by screening out variables at the beginning of each $\lambda$ iteration.  The form of the sequential strong rules is described at the end of the section.

\subsection{Blockwise descent (squared-error loss)}  \label{sec:descent_linear}
This section describes the update steps for the blockwise coordinate descent algorithm.  Optimizing over $\alpha_j$ for a fixed $j$ given the current values of the other parameters amounts to a simple one-dimensional lasso problem whose solution is given in closed form by soft thresholding.  The optimization step for $\beta_j$ is rather more involved, and we find that we are unable to obtain an entirely closed form solution to the corresponding subgradient equation.  We instead give a closed form solution up to a quantity that is calculated via a simple one-dimensional line search.  

In deriving the update steps we assume that the predictors have been centred and normalized.


\textit{Intercept term.} Since the predictors are centred and the intercept term is left unpenalized in the objective, the intercept estimate is given by
\[
 \hat \alpha_0 = \frac{1}{n}\sum_{i=1}^n y_i
\] 


\textit{Optimization in $\alpha$.}  Given current estimates $\tilde \alpha_k$ for $k\neq j$ and $\tilde \beta_k$ for all $k$,  let $r^{(j)}_\alpha$ denote the residual vector 
\[
r_\alpha^{(j)} =y - \left(\hat \alpha_0 + \sum_{k \neq j} \tilde \alpha_k x_k + \sum_{k=1}^n U_k \tilde\beta_k \right)
\]

With this notation, the optimization problem \eqref{eq:objective} reduces to a simple one-dimensional lasso problem,
\[
  \hat \alpha_j  = \argmin_{\alpha_j} \, \frac12\| r^{(j)}_\alpha - \alpha_j x_j\|_2^2 + \gamma\lambda |\alpha_j|
\]
The solution to this is known to be given by,
\begin{equation}
\hat \alpha_j = S(x^T_j r^{(j)}_\alpha; \gamma\lambda) \label{eq:alphastep}
\end{equation}
where  $S(y;t)$ is the soft-thresholding function defined as $S(x;\lambda)=\sign(x)(|x|-\lambda)_+$.

\textit{Optimization in $\beta$.}  Given current estimates $\tilde \beta_k$ for $k\neq j$ and $\tilde \alpha_k$ for all $k$,  let $r^{(j)}_\beta$ denote the residual vector 
\[
r_\beta^{(j)} =y - \left(\hat \alpha_0 + \sum_{k=1}^n \tilde \alpha_k x_k + \sum_{k\neq j} U_k \tilde\beta_k \right)
\]
As a function of $\beta_j$, the optimization problem \eqref{eq:objective} reduces to
\[ 
\tilde \beta_j = \argmin_{\beta_j} \frac{1}{2}\left\| r^{(j)}_\beta -U_j\beta_j \right\|^2
  +\lambda(1 - \gamma) {\trdinv} \|\beta_j\|_{D_j^*} + \psi_j \beta_j^T D_j \beta_j
\]
In order to simplify notation we temporarily drop the explicit dependence on $j$ and make the substitutions $\tilde \lambda = \lambda(1-\gamma){\trdinv}$, $\theta = {D^*}^{1/2}\beta_j$, $\theta_{(-1)} = (0, \theta_2, \ldots, \theta_{m_j})$ and $V = U_j{D^*}^{-1/2}$ so that the above equation becomes

\[ 
\hat \theta = \argmin_{\theta} \frac{1}{2}\left\| r -V\theta \right\|^2
  +\tilde\lambda\|\theta\|_2 + \frac{1}{2}\psi \theta_{(-1)}^T\theta_{(-1)}
\]
Differentiating the objective with respect to $\theta$ gives the subgradient equation,
\begin{equation}
  -V^T(r - V\theta)+\tilde\lambda\subgrad + \psi \theta_{(-1)} = 0
  \label{eq:theta_subgrad}
\end{equation}
where $\subgrad$ is in the subgradient of the $\ell_2$ norm: 
\begin{align*}
  &\subgrad \in \{ u \in \R^{m_j} : \|u\|_2 \le 1 \} &\text{ if } \theta \equiv 0 \\
  &\subgrad = \frac{\theta}{\|\theta\|_2} &\text{ otherwise}
\end{align*}

From the subgradient equation \eqref{eq:theta_subgrad}, we see that $\theta \equiv 0$ is a solution if $\|V^Tr\|_2 \le \tilde\lambda$.  Otherwise, the subgradient equation becomes
\[
-V^T(r - V\theta) + \psi\theta_{(-1)} + \tilde\lambda \frac{\theta}{\|\theta\|_2} = 0
\]
Some algebraic manipulation along with the substitution $\widetilde{D} = V^TV + \psi I_{(-1)} = {D^*}^{-1} + \psi I_{(-1)}$ produces the simpler expression,
\begin{equation}
  \theta =\left( \widetilde{D} + \frac{\tilde\lambda}{\|\theta\|_2}I \right)^{-1} V^Tr
\label{eq:thetasoln}
\end{equation}

If the matrix $\widetilde{D}$ were of the form $\widetilde{D} = c I$, then this equation would have a simple closed form solution for $\theta$.  This condition is typically not satisfied, so we proceed by first solving for $c=\|\theta\|_2$ and then substituting this quantity back into \eqref{eq:thetasoln} to solve for $\theta$.

Taking the squared norm of both sides of \eqref{eq:thetasoln} and doing some algebra gives
\begin{equation}
  \sum_{i=1}^{m_j} \left(\frac{(V^Tr)_i}{\widetilde{D}_i c + \tilde\lambda}\right)^2 - 1 = 0
  \label{eq:linesearch}
\end{equation}
Solving for $c$ amounts to carrying out a simple one-dimensional line search.  

Let $\hat c$ denote the solution to \eqref{eq:linesearch}.  Reversing all the substitutions we made along the way, we conclude that the update step for $\beta_j$ looks like

\begin{equation}
  \hat \beta_j = \begin{cases}
  0, \qquad \text{if } \|{D^*}^{-1/2}U_j^T r^{(j)}_\beta \|_2 \le \lambda(1-\gamma){\trdinv} \\
  \left({D^*}^{-1} + \psi_j I_{(-1)} + \frac{\lambda(1-\gamma){\trdinv}}{\hat{c}}I  \right)^{-1} {D^*}^{-1}U_j^T r^{(j)}_\beta, \qquad \text{otherwise}
\end{cases}
  \label{eq:betastep}
\end{equation}

\subsection{Logistic regression} \label{sec:logistic}
In this section we describe the blockwise gradient descent algorithm that we use for logistic regression.  The log-likelihood function for logistic regression has the form, 

\begin{equation}
\ell(\alpha_0, \alpha, \beta)  = \sum_{i=1}^n y_i\left(\alpha_0 + \sum_{j=1}^p f_j(x_i)\right) - \log\left(1 + e^{ \alpha_0 + \sum_{j=1}^p f_j(x_i) }\right)
\label{eq:logistic_loss}
\end{equation}

A common approach is to use a Taylor expansion about the current parameter estimates to obtain a quadratic approximation to the log-likelihood.  This results in a weighted residual sum of squares expression of the form,
\begin{equation}
  \ell_Q(\alpha_0, \alpha, \beta) = \frac{1}{2}\sum_{i=1}^n w_i\left(z_i - \alpha_0 - \sum_{j=1}^p f_j(x_i) \right)^2 + C
  \label{eq:quad_approx}
\end{equation}
where 
\begin{align}
 z_i &= \tilde \alpha_0 + \sum_{j=1}^p \tilde f_j(x_i)  + \frac{y_i - \tilde p(x_i)}{w_i} &\text{(working response)}\\
 w_i &= \tilde p(x_i)(1- \tilde p(x_i))  &\text{(weights)} \\
 \tilde p(x_i) &= \left[ 1 + e^{-\left( \tilde \alpha_0 + \sum_{j=1}^n \tilde f_j(x_i) \right)} \right]^{-1}
\end{align} 
See \cite{hastie1990generalized} for details.

The full penalized weighted least squares objective then has the form
\begin{multline}
  \label{eq:logistic_objective}
\frac{1}{2} \left\| W^{1/2} \left(z - \alpha_0 - \sum_{j = 1}^p \alpha_j x_j - \sum_{j = 1}^p U_j \beta_j \right)\right\|_2^2
\\
 +  \lambda\sum_{j = 1}^p\left( \gamma |\alpha_j| + (1-\gamma){\trdinv} \|\beta_j\|_{D^*}\right) + \frac{1}{2}\sum_{j = 1}^p \psi_j \beta_{j}^T D \beta_{j} 
\end{multline}

It turns out that, for general values of the weights $w_i$, the update step for the $\beta_j$ terms amounts to solving an equation having the same form as \eqref{eq:linesearch}, with the added difficulty that now $\widetilde{D}$ is non-diagonal.  While the resulting equation is possible to solve (say, by evaluating the eigen-decomposition of $\widetilde{D}$), doing so incurs a greater computational cost than incurred in the linear regression setting.  

Our approach is to instead use a coarser majorization, replacing all of the weights with $w_i = 0.25$, the maximum possible.  This reduces \eqref{eq:logistic_objective} to the same objective as in the linear regression case, with response given by the working response vector, $z$.  In summary, we proceed as follows.

\begin{description}
\item[\it Outer loop:] Decrement $\lambda$. 
\item[\it Middle loop:] Update the quadratic approximation $\ell_Q$ using the current values of the parameters, $\{\tilde\alpha_0, \alpha_j, \beta_j\}_{j=1}^p$, setting $w_i = 0.25$. 
\item[\it Inner loop:] Run the coordinate descent procedure of \textsection\ref{sec:descent_linear} on the penalized least squares problem \eqref{eq:logistic_objective} until convergence.
\end{description}

\subsection{Strong rules}  Strong rules as introduced in \cite{tibshirani2012strong} are an effective computational hedge for discarding inactive predictors along the $\lambda$ path.  Given the solution at $\lambda_{k-1}$, strong rules give a simple check for screening out predictors that are likely to be inactive in the solution at $\lambda_k < \lambda_{k-1}$.
The idea is to ignore these discarded variables, fit the model, and then confirm their omission. If any are omitted in error, they are then included in the active set, and the fitting is repeated. In practice, this refitting is very rarely needed.

Letting $\hat y(\lambda)$ denote the fitted values at penalty parameter value $\lambda$ (probabilities in the case of logistic regression), the sequential strong rules for GAMSEL can be shown to be

\begin{description}
  \item[\it Strong rule for $\alpha_j$:] Discard $\alpha_j$ if
  \[
    \left|x_j^T(y-\hat y(\lambda_{k-1})) \right| < \gamma(2\lambda_k - \lambda_{k-1})
  \]
  \item[\it Strong rule for $\beta_j$:] Discard $\beta_j$ if
  \[
    \left\| U_j^T\left(y - \hat y(\lambda_{k-1})\right) + \psi_j D\beta_j  \right\|_2
    < (1-\gamma)(2\lambda_k - \lambda_{k-1})
  \]
\end{description}

In the implementation of GAMSEL we adopt the approach recommended by the authors in section 7 of \cite{tibshirani2012strong}, which takes advantage of both active set methods and strong rules screening.

These strong-rule formulas also expose to us the the value of $\lambda_{max}$, the largest value of $\lambda$ we need to consider in the GAMSEL regularization path:
\begin{equation}
  \label{eq:lambdamax}
  \lambda_{max}=\max\left\{\max_j\frac{|x_j^Ty|}\gamma,\;\max_j\frac{\|U_j^Ty\|_2}{1-\gamma}\right\}.
\end{equation}
For any values of $\lambda>\lambda_{max}$, all terms in the model  bar the intercept are zero.


\section{Experiments}\label{sec:experiments}
We now demonstrate the GAMSEL procedure on synthetic and real data.
\subsection{Simulation study} \label{sec:simstudy}
 We begin this section with a simple simulation where we have $n=200$ observations on $p=30$.  Variables 1--6 are linear; 7--10 are polynomials of degree $5$, and the remaining 20 variables (indexes 11--30) are zero. All coefficients in the simulation are generated randomly.  We begin by running GAMSEL with $\gamma=0.4$ and use 10 basis functions with $5$ degrees of freedom for each of the $30$ variables.  
 
\begin{figure}[!ht]
    \centering
    \begin{minipage}{0.03\textwidth}
      $\lambda_5$
    \end{minipage} 
    \begin{minipage}{0.96\textwidth}
      \includegraphics[width=\textwidth]{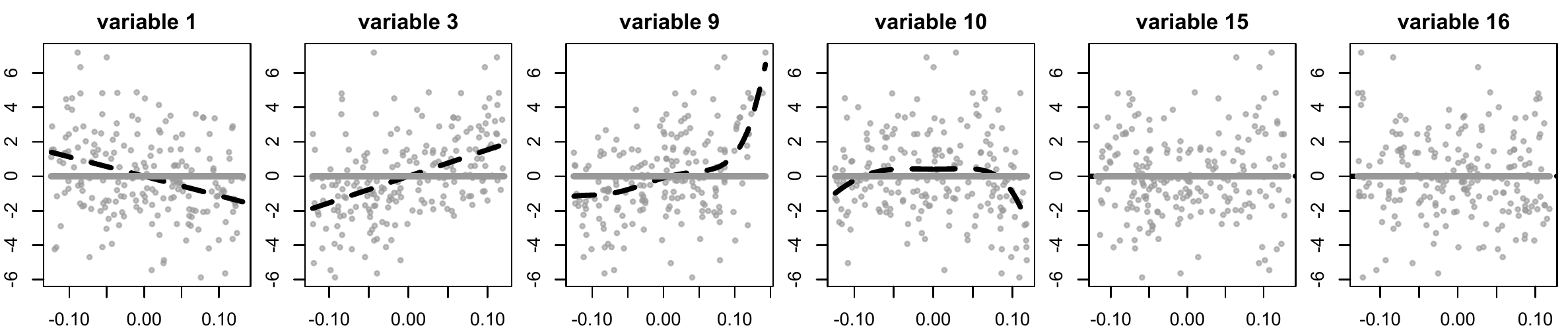}
    \end{minipage}
    \begin{minipage}{0.03\textwidth}
      $\lambda_{15}$
    \end{minipage} 
    \begin{minipage}{0.96\textwidth}
      \includegraphics[width=\textwidth]{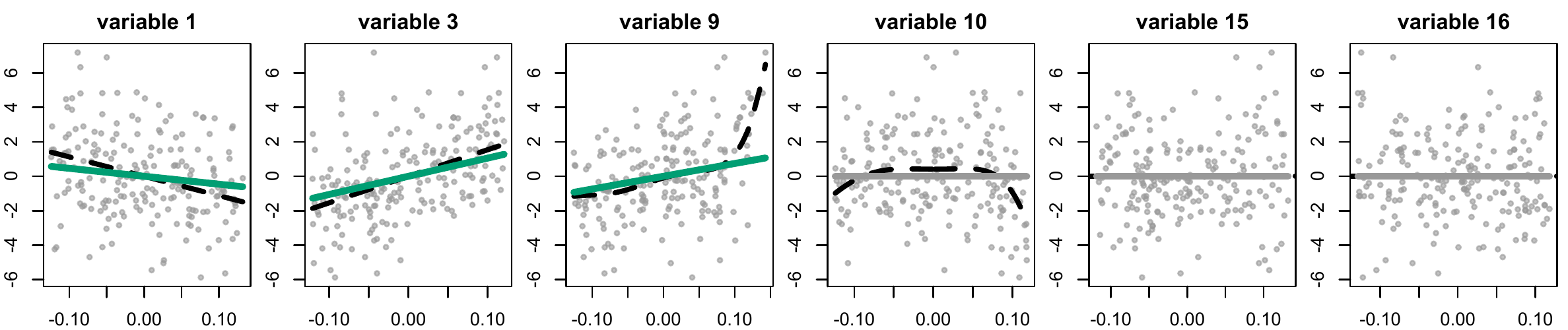}
    \end{minipage}
    \begin{minipage}{0.03\textwidth}
      $\lambda_{25}$
    \end{minipage} 
    \begin{minipage}{0.96\textwidth}
      \includegraphics[width=\textwidth]{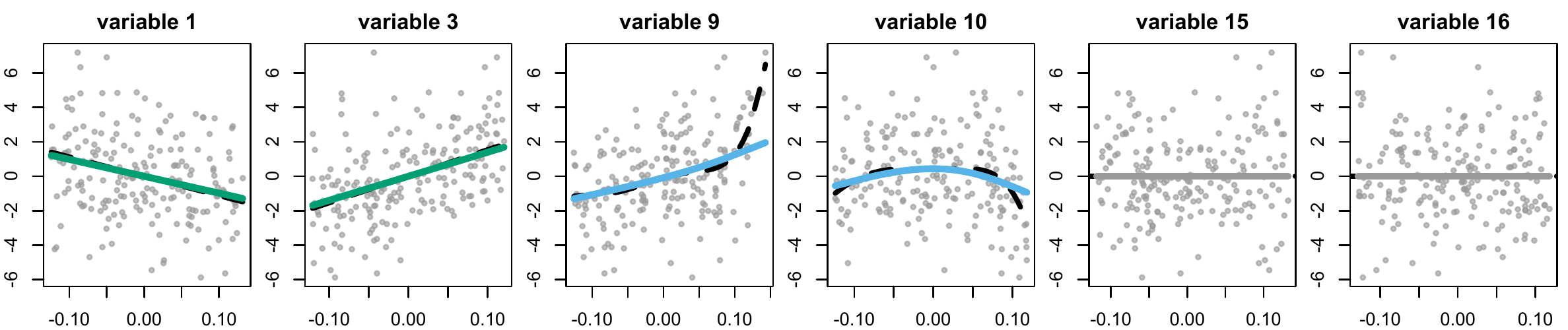}
    \end{minipage}
    \begin{minipage}{0.03\textwidth}
      $\lambda_{40}$
    \end{minipage} 
    \begin{minipage}{0.96\textwidth}
      \includegraphics[width=\textwidth]{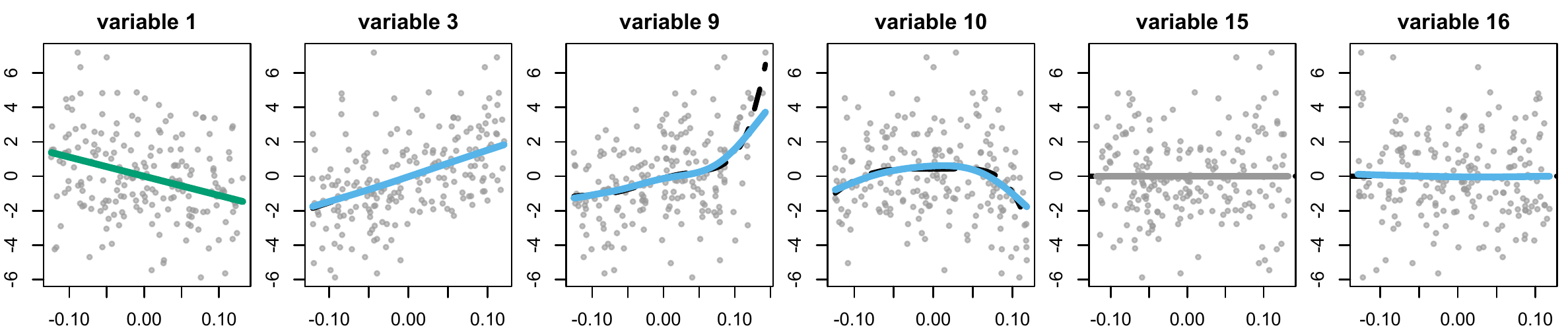}
    \end{minipage}
    \caption{Estimated $f_j$ for 6 variables across four $\lambda$ values for a single realization of the simulation described in \ref{sec:simstudy}.  Linear fits are displayed in green; nonlinear fits are displayed in blue. At $\lambda_{25}$ all 6 variables are correctly identified as linear, non-linear and zero.}
    \label{fig:simfits}
\end{figure}

 Figure~\ref{fig:simfits} shows plots of the fitted coefficients of 6 variables for several values of the penalty parameter $\lambda$.   At $\lambda_{25}$ all six variables shown are correctly classified as linear, nonlinear or zero.  By $\lambda_{40}$, variable 3 gets a non-linear fit, as does the noise variable at index 16.  The estimated functions for variables $9$ and $10$ become increasingly better matches for the underlying signals as $\lambda$ decreases.  

\begin{figure}[!ht]
    \centering
    \begin{subfigure}[b]{0.32\textwidth}
    \includegraphics[width=\textwidth]{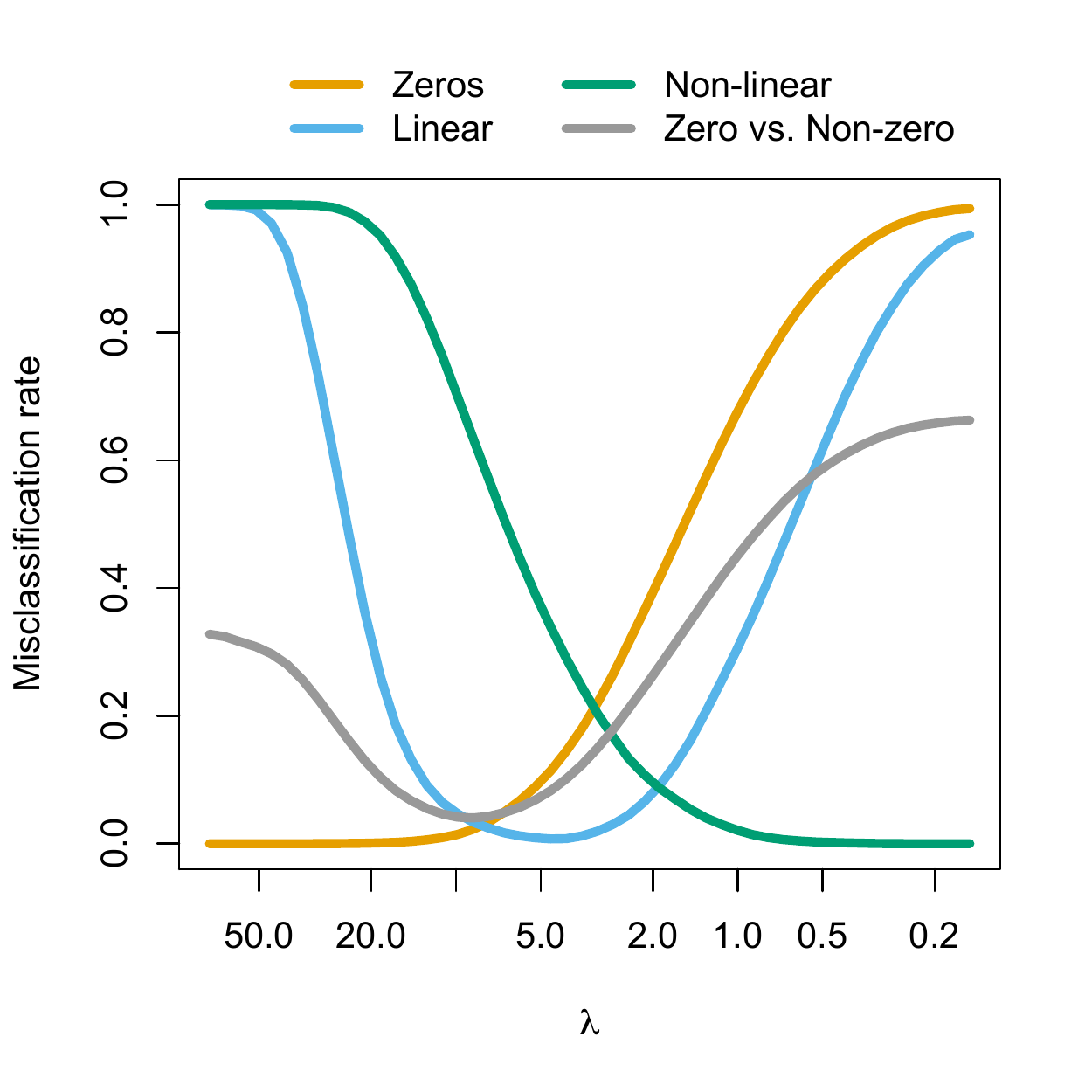}
    \caption{$\gamma=0.4$}
    \end{subfigure}
    \begin{subfigure}[b]{0.32\textwidth}
    \includegraphics[width=\textwidth]{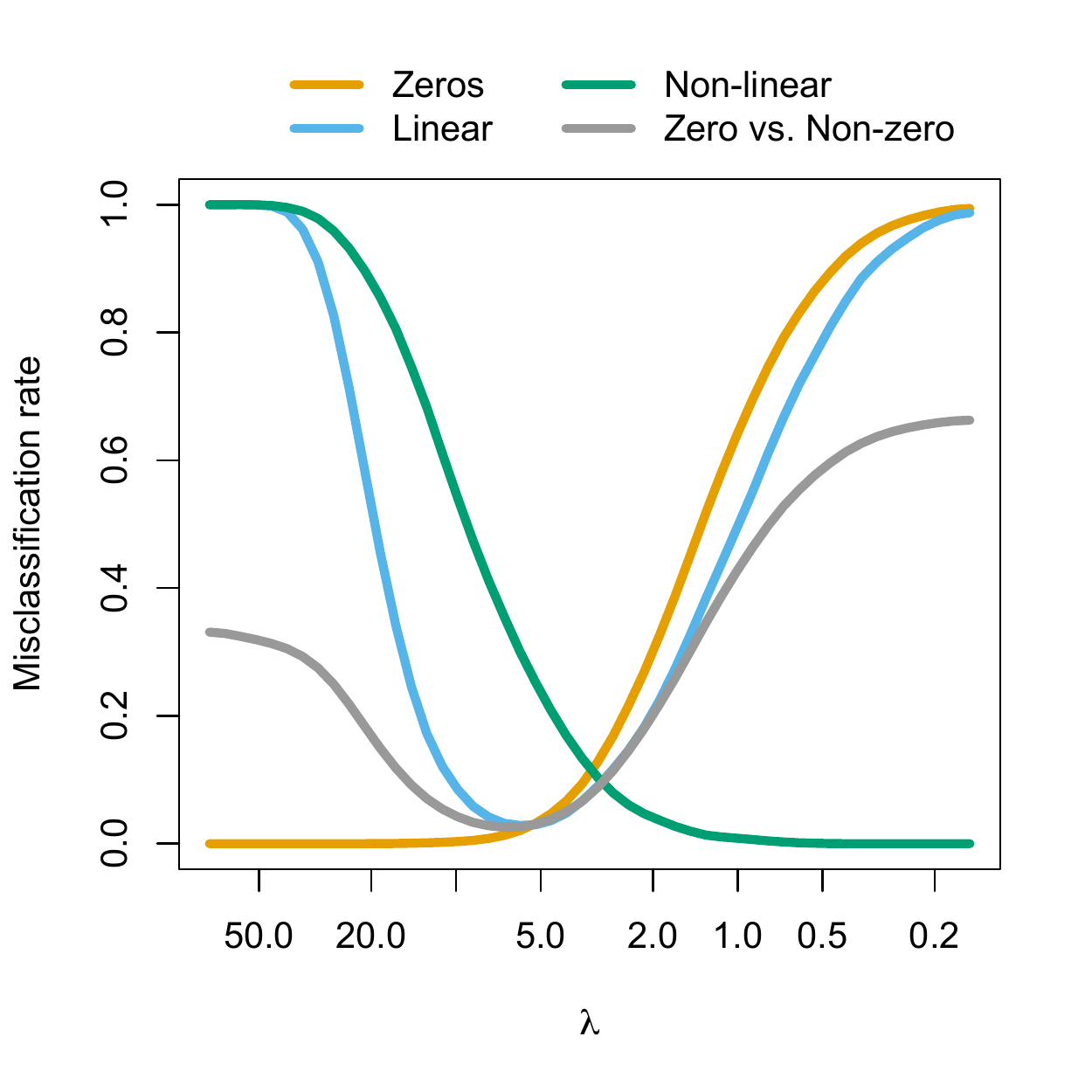}
    \caption{$\gamma=0.5$}
    \end{subfigure}
    \begin{subfigure}[b]{0.32\textwidth}
    \includegraphics[width=\textwidth]{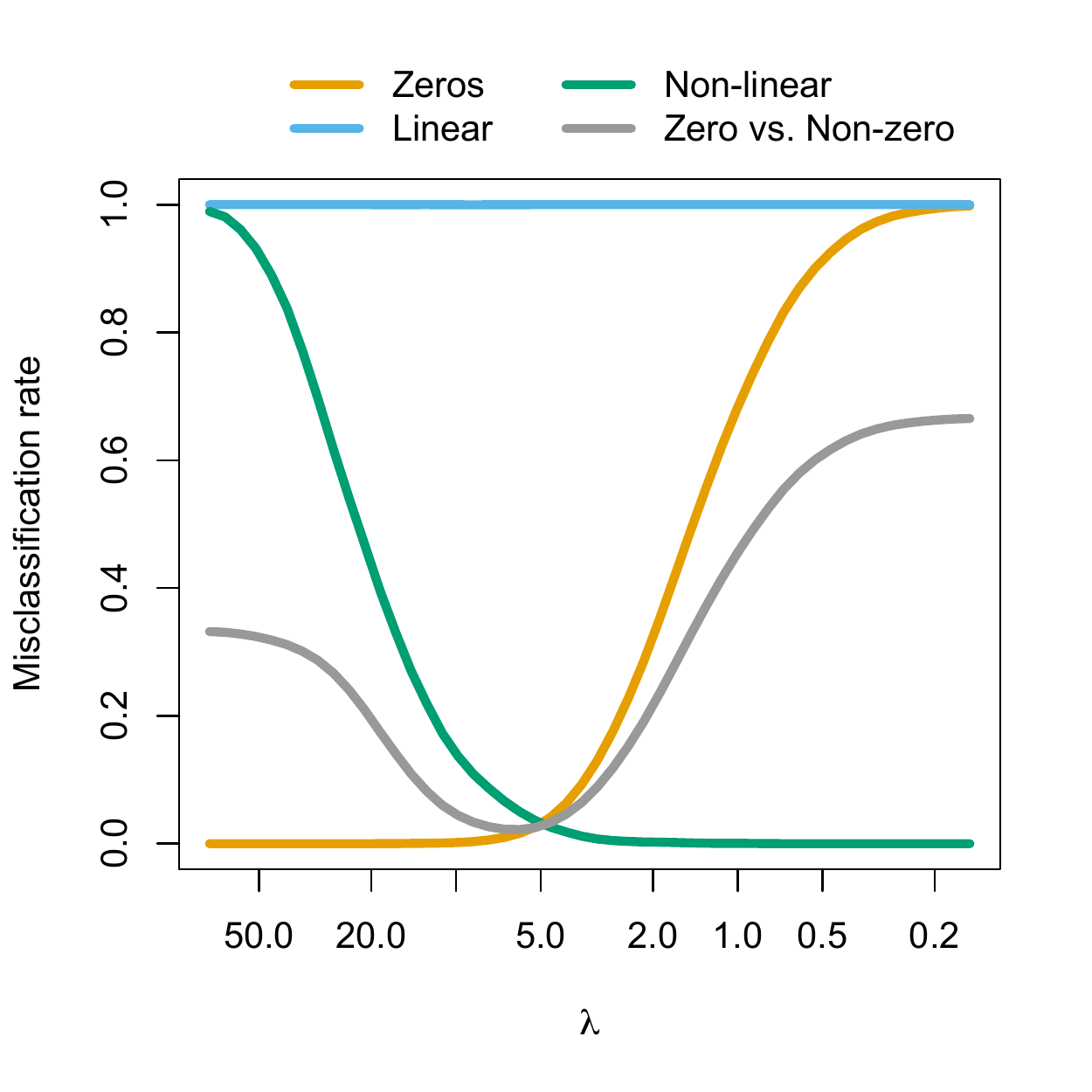}
    \caption{$\gamma=0.6$}
    \end{subfigure}
    \caption{This figure illustrates the classification performance of GAMSEL on the four misclassification measures described in section \ref{sec:simstudy}.   We observe that the overall Zero vs Non-zero misclassification curve is not sensitive to the choice of tuning parameter $\gamma$.  As we vary $\gamma$, the main difference is in how the terms are classified between linear and non-linear.  Since non-linear terms can be used to represent linear ones, it is in general inadvisable to take $\gamma > 0.5$.}
    \label{fig:simclass}
\end{figure}

We repeat the simulation $500$ times at three values of $\gamma$: 0.4, 0.5, 0.6.  Figure~\ref{fig:simclass} shows plots of misclassification error across the full sequence of $\lambda$ values for the three choices of $\gamma$.  Four types of misclassification are considered:
\begin{description}
  \item[\it Zeros.] The fraction of truly zero variables that are estimated as non-zero (i.e., as linear or non-linear).
  \item[\it Linear.]  The fraction of truly linear variables that are estimated as zero or non-linear.
  \item[\it Non-linear.] The fraction of truly non-linear variables that are estimated as zero or linear.
  \item[\it Zero vs. Non-zero.] The misclassification rate between the categories zero and non-zero.  
\end{description}

We see from the Figure~\ref{fig:simclass} that the Zero vs. Non-zero misclassification rate as well as the Zeros misclassification rate are fairly insensitive to the choice of $\gamma$.  By design, Linear and Non-linear misclassifications are very sensitive to the choice of $\gamma$.  Since non-linear fits can be used to represent linear functions but not vice-versa, taking $\gamma$ much larger than $0.5$ will result in all nonzero fits being non-linear.  This behaviour is observed in the panel corresponding to $\gamma=0.6$ in the present simulation.

\subsection{Boston Housing}

\begin{figure}[!ht]
  \centering
    \includegraphics[width=0.75\textwidth]{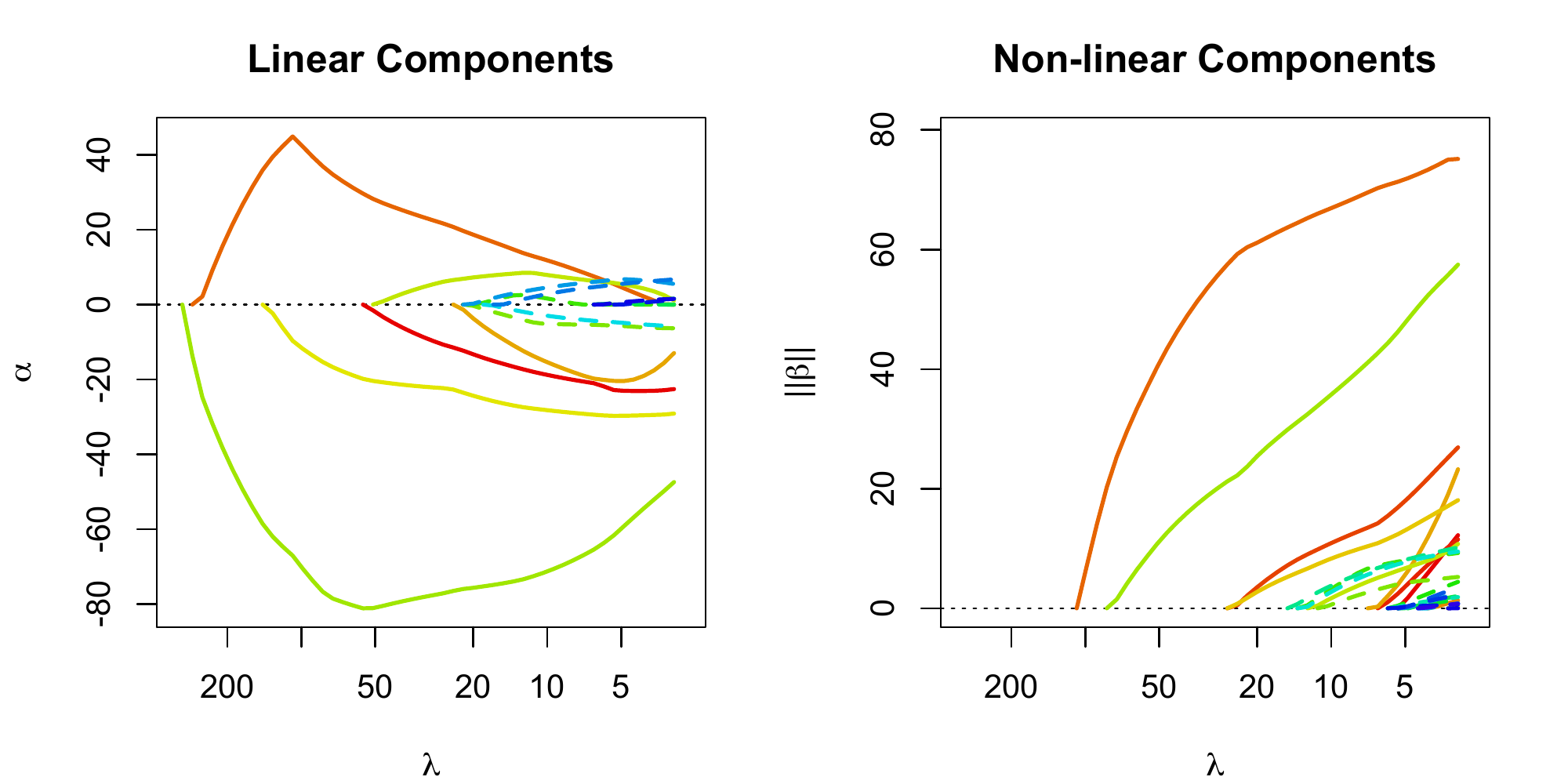}
  \caption{Regularization plots for the Boston Housing data.  Solid lines correspond to the original 10 variables; dashed lines are from the simulated noise variables.}
  \label{fig:bostonreg}
\end{figure}
The Boston Housing Dataset contains housing information on median house values for $n=506$ census tracts in the Boston Standard Metropolitan Statistical Area.  Of the $13$ covariates present in the dataset, $p=10$ are sufficiently continuous to be meaningfully analysed within the GAMSEL framework.  These variables are: CRIME, INDUS, NOX, RM, AGE, TAX, PTRATIO, B, and LSTAT.  In addition to considering the original variables in the dataset, we follow the approach of \cite{ravikumar2007spam} and generate $20$ noise variables to include in the analysis.  The first ten are drawn uniformly from the interval $[0,1]$, while the remaining ten are obtained as random permutations of the original ten covariates.  
\begin{figure}[!ht]
  \centering
  \includegraphics[width=0.4\textwidth]{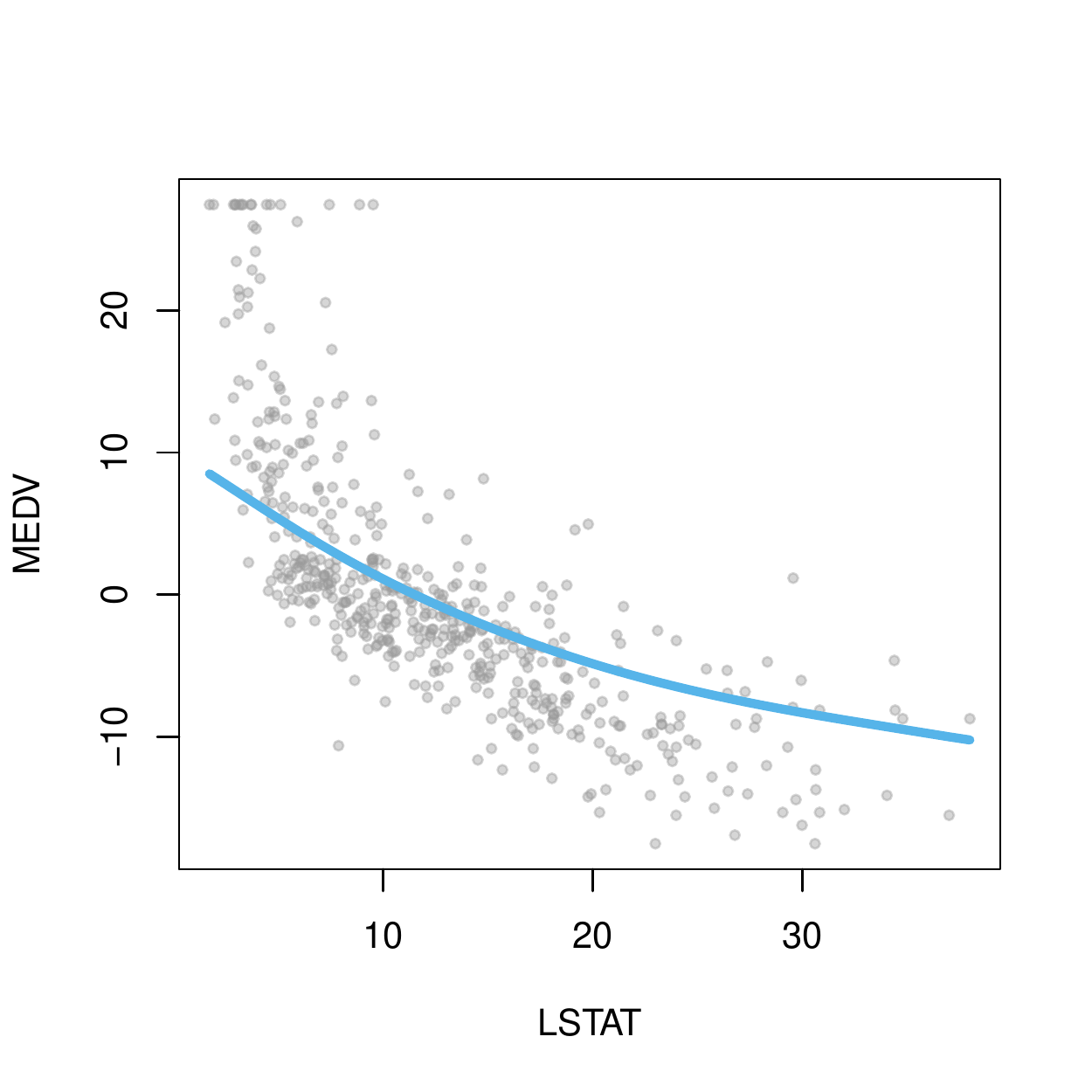} 
  \includegraphics[width=0.4\textwidth]{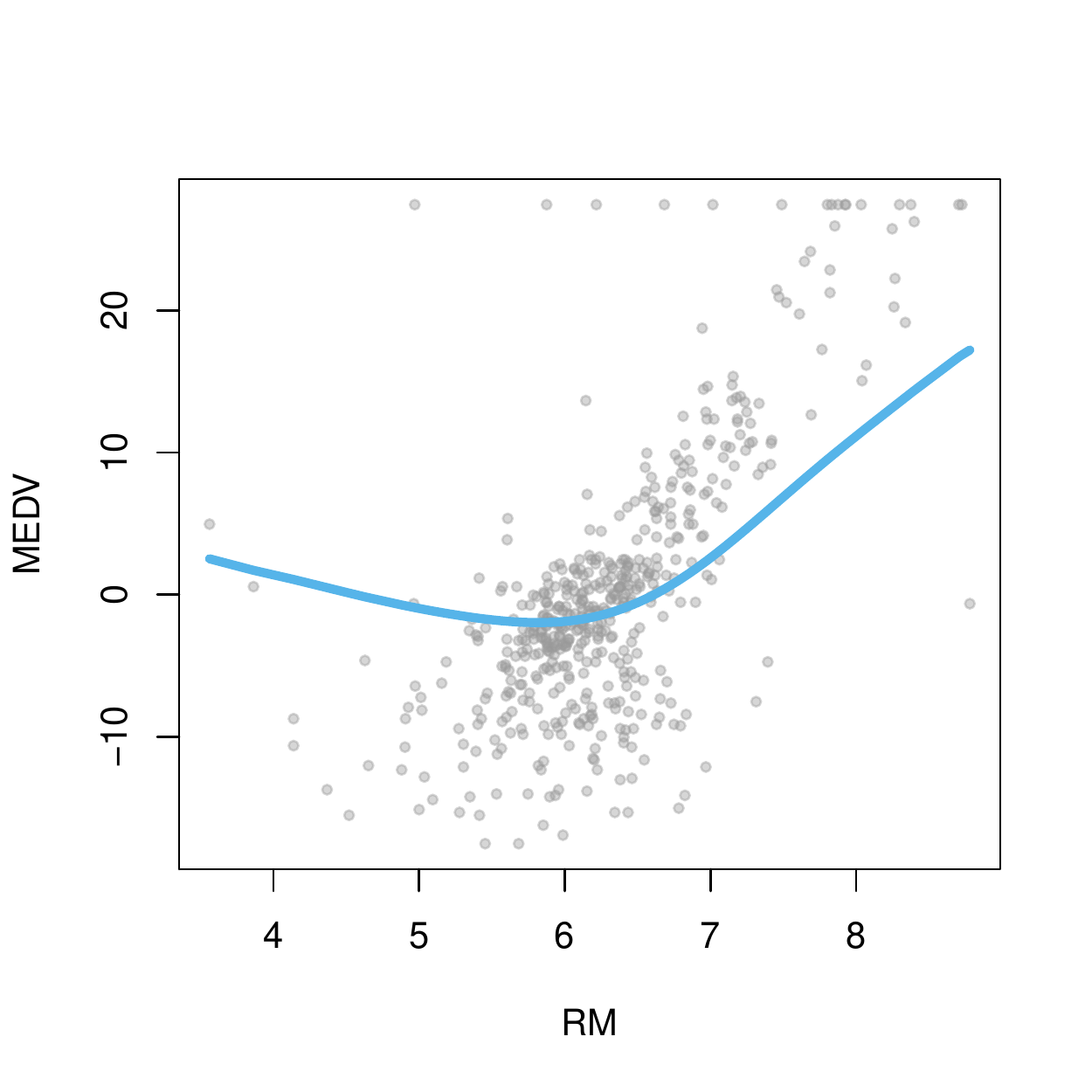} \\[-30pt]
  \includegraphics[width=0.4\textwidth]{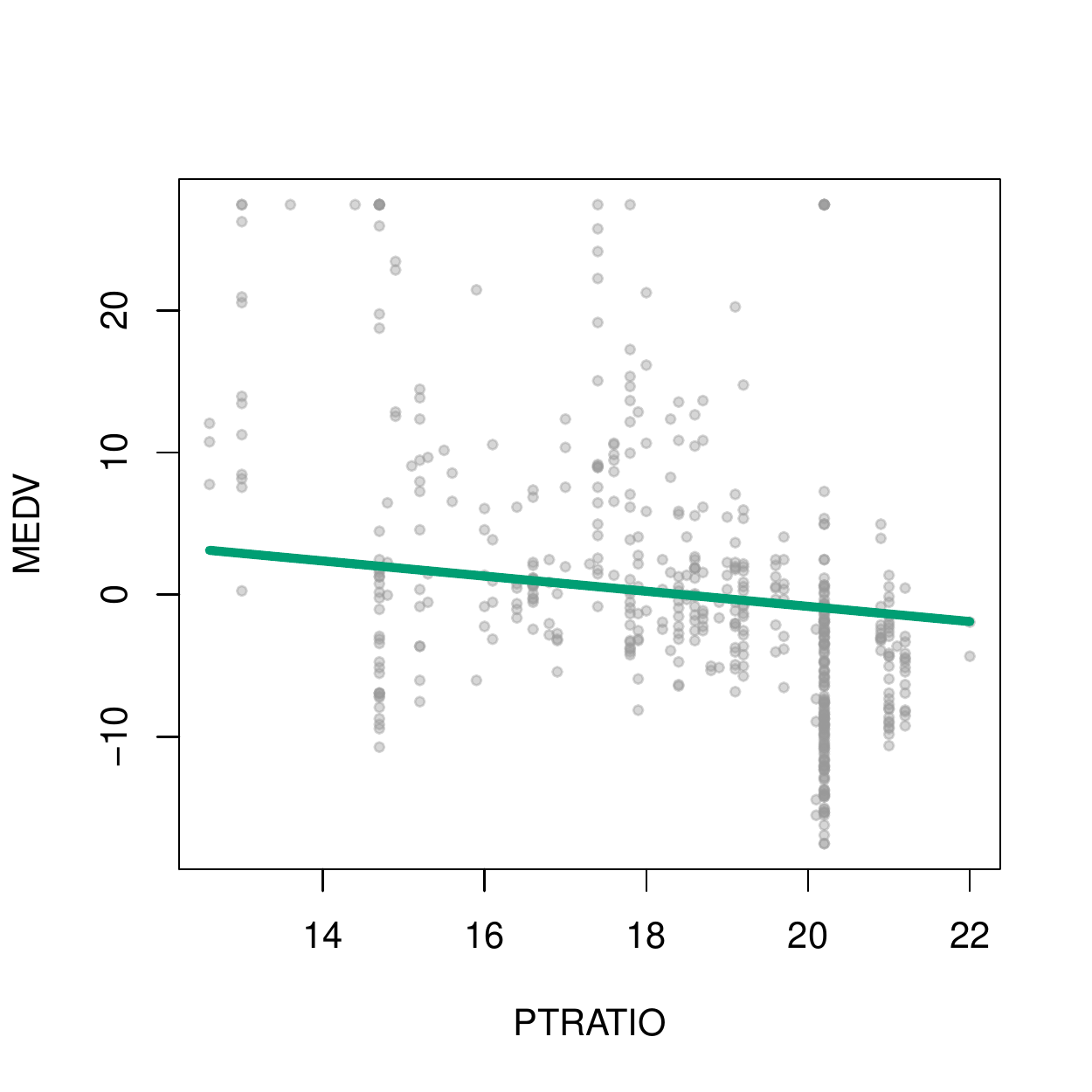} 
  \includegraphics[width=0.4\textwidth]{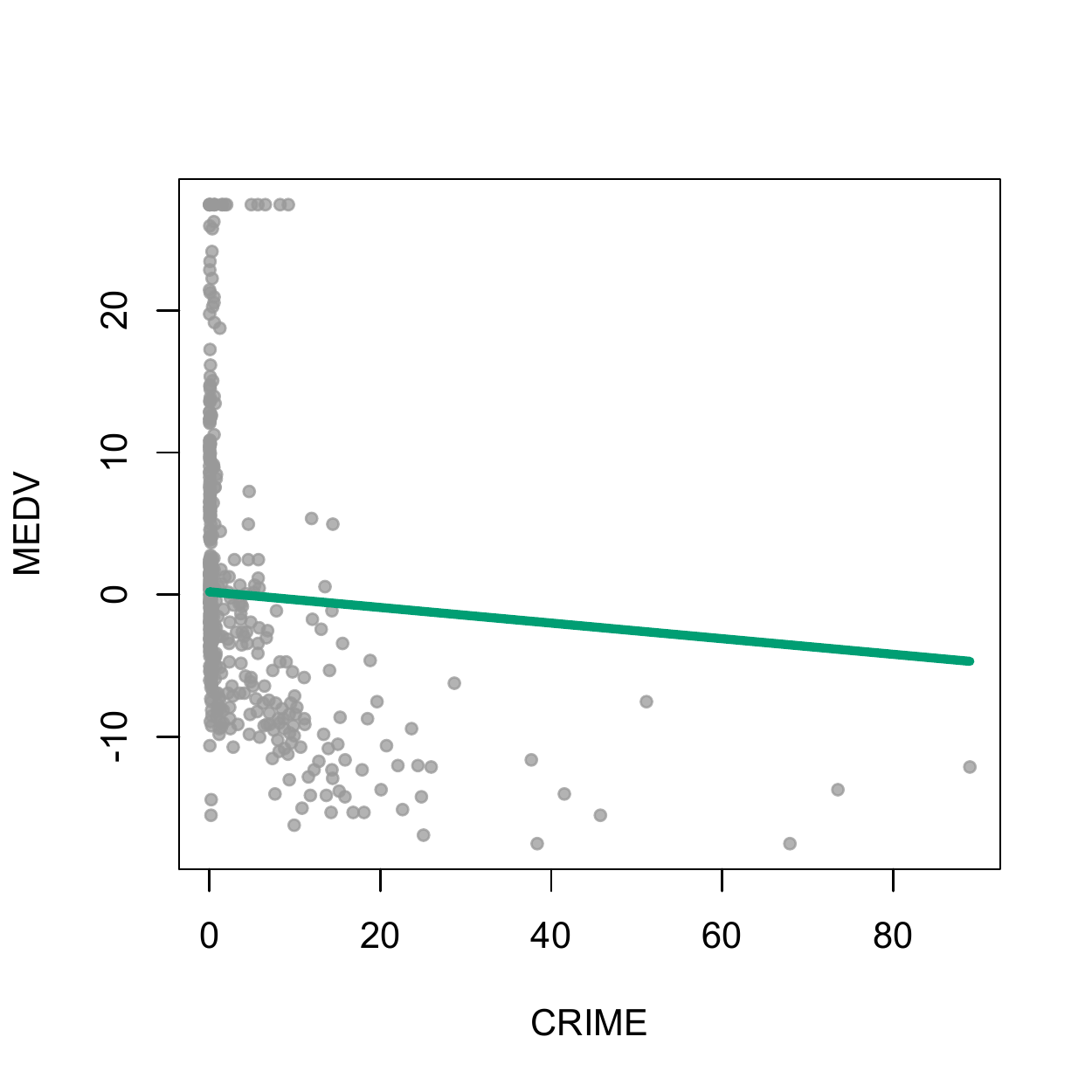}
  \caption{Boston Housing: Estimated functions at $\lambda_{28} = 24.1$.  }
  \label{fig:bostonfit}
\end{figure}
We fit GAMSEL with $\gamma=0.5$ using $10$ basis functions with $5$ degrees of freedom for all $30$ covariates.  Figure~\ref{fig:bostonreg} summarizes the regularization plots, which display $\alpha_j$ and $\|\beta_j\|_2$ across the sequence of $\lambda$ values.  This plot shows that there are $5$ strong predictors that enter well before the `bulk': LSTAT (nonlinear), RM (nonlinear), PTRATIO (linear), CRIME (linear), and B (linear).  TAX and NOX also enter the model before any of the noise variables, but they are somewhat borderline.   Figure~\ref{fig:bostonfit} shows estimated functions for LSTAT, RM, PTRATIO and CRIME at $\lambda_{29}=24.1$, right before the noise variables enter.  The fitted curves do a good job of capturing the trends in the data.

\subsection{Spam Data} \label{sec:spamdata}

Next we apply GAMSEL to a spam classification problem using the Hewlett-Packard spam dataset.  This dataset consists of textual information from $n=4601$ email messages, each of which is labelled \emph{spam} or \emph{email}.  There are $57$ predictors, of which $48$ give the percentage of words matching a given word (e.g., `free', `hp', `george'); 6 give the percentage of characters that match a given character (e.g., \$, !); and 3 measure uninterrupted sequences of capital letters. We conduct two analyses of the spam data.  
\begin{figure}[!ht]
  \centering
  \begin{minipage}{0.59\textwidth}
    \includegraphics[width=\textwidth]{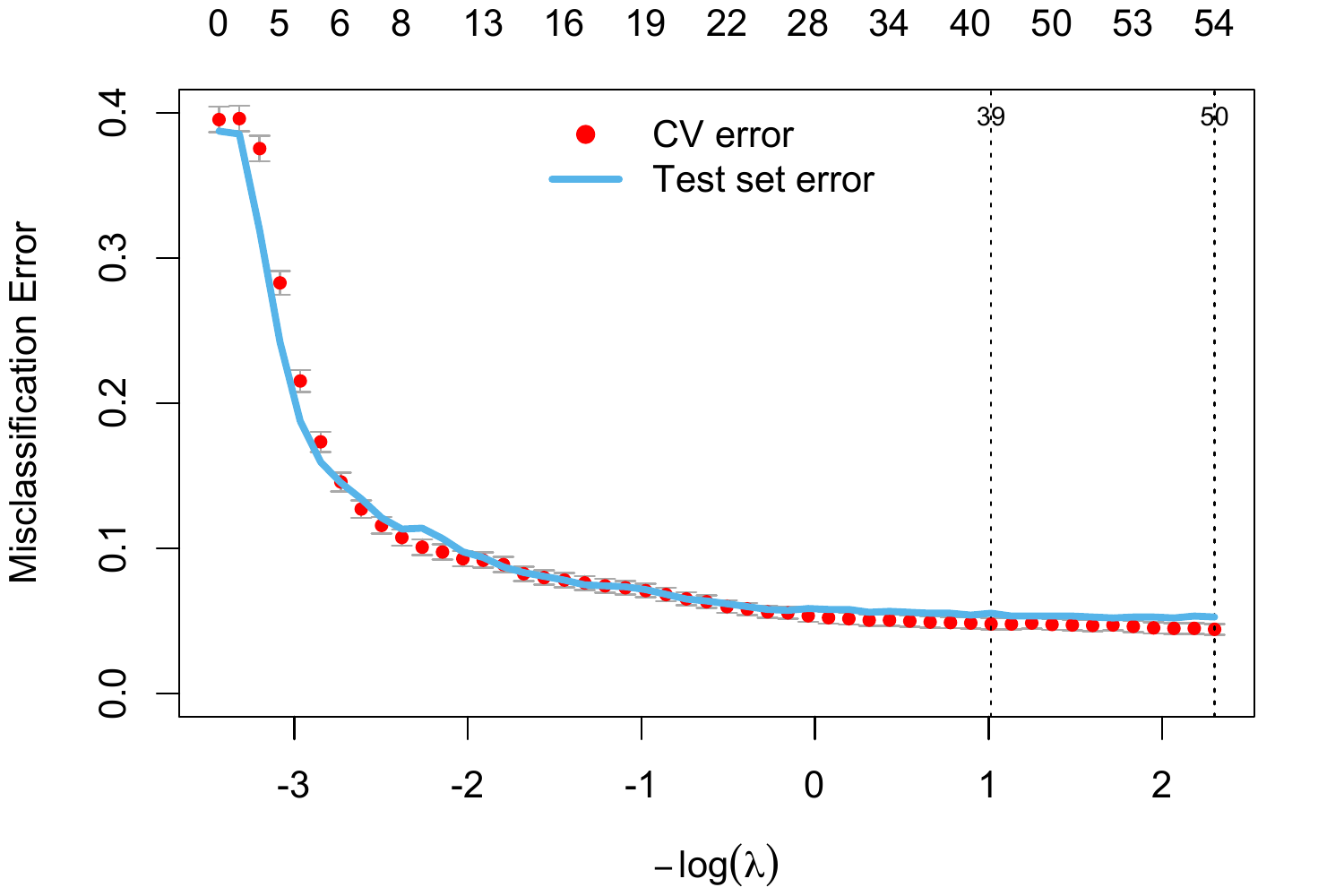}
  \end{minipage}
  \begin{minipage}{0.39\textwidth}
      \includegraphics[width=\textwidth]{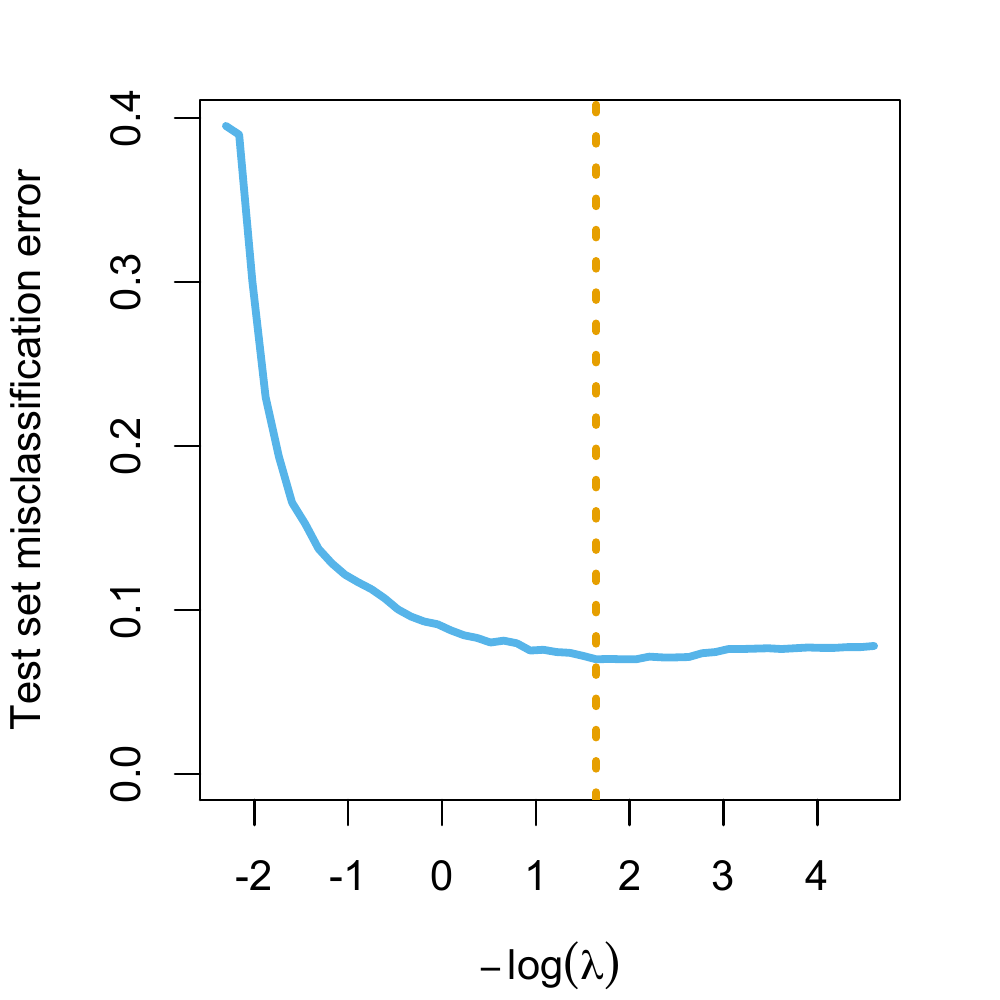}
  \end{minipage}
  \caption{Minimum test error is $7.0\%$ for the plot shown. 1se $\lambda$ from CV gave a test error of $5.5\%$}
  \label{fig:spamdata}
\end{figure}
First, we analyse the data using the test set and training set partitioning described in section 9.1.2 of \cite{hastie2009elements}.  Using the 3065 observations in the training set, we fit GAMSEL with $\gamma=0.5$, taking $10$ basis functions with $4$ degrees of freedom for all $57$ covariates.   The tuning parameter $\lambda$ is chosen via cross-validation using the 1 standard error rule.  The left panel of Figure~\ref{fig:spamdata} shows the full cross-validation curve along with the test set misclassification error curve across the entire sequence of $\lambda$ values.  We see that the one-standard-error rule selects the model corresponding to $\lambda_{39}=0.362$.  The selected model has cross-validated misclassification error of $4.8\%$, and test misclassification error of $5.5\%$.   This compares favourably with test set error rate of $5.3\%$ reported in \cite{hastie2009elements}, as found by \texttt{step.gam}.

In our second analysis we further sub-sample the data to obtain a training sample of $n=300$ messages, with the remaining $4301$ messages used for testing.  Due to the extreme skewness in some of the covariates, the training sample contained several variables for which as few as $3$ unique values were observed.  For the purpose of this analysis we used (degree, df) $\in \{(10,4), (4,2), (1,1)  \}$, as determined by the number of unique values observed in the test sample for the given variable.  The right plot of Figure~\ref{fig:spamdata} shows the test set misclassification error plotted against the penalty parameter $\lambda$ for a single realization of the training set.  The minimum achieved test error is $7.0\%$.

\section{Comparisons to other methods}\label{sec:comp-other-meth}

In this section we compare GAMSEL with three other approaches to carrying out selection and estimation for generalized additive models:
\begin{enumerate}
\item  the SpAM method of \cite{ravikumar2007spam},
\item the {\tt step.gam} routine in the {\tt gam} package, and
\item the {\tt gam.selection} procedure in the {\tt mgcv} package.
\end{enumerate}
The SpAM method does not separate linear from nonlinear terms, but instead identifies terms to be set to zero. It is a modified form of backfitting, where each fitted function is thresholded via a group-lasso penalty each time it is updated. See \citet[Chapter 4]{hastie15:_statis_learn_with_spars} for a succinct description.  

\subsection{GAMSEL vs. SpAM}

We begin by comparing the model selection performance of GAMSEL and SpAM in the simulation setting described in \ref{sec:simstudy}.  As before, we have $n=200$ observations on $p=30$ variables, of which $10$ are non-zero.  We considered three simulation sub-settings: \#(linear, nonlinear) $=$ (a) $(10,0)$ (b)$(6,4)$, (c)$(0,10)$.  We ran GAMSEL with $10$ basis functions and $5$ degrees of freedom and $\gamma = 0.5$, and using the {\tt SAM} package we ran SpAM with the default $k=3$ basis spline functions.  In our experimentation we also explored taking $k > 3$, but we observed that this resulted in a deterioration in the model selection performance of SpAM. 
\begin{figure}[!ht]
  \centering
  \begin{subfigure}[b]{0.32\textwidth}
  \includegraphics[width=\textwidth]{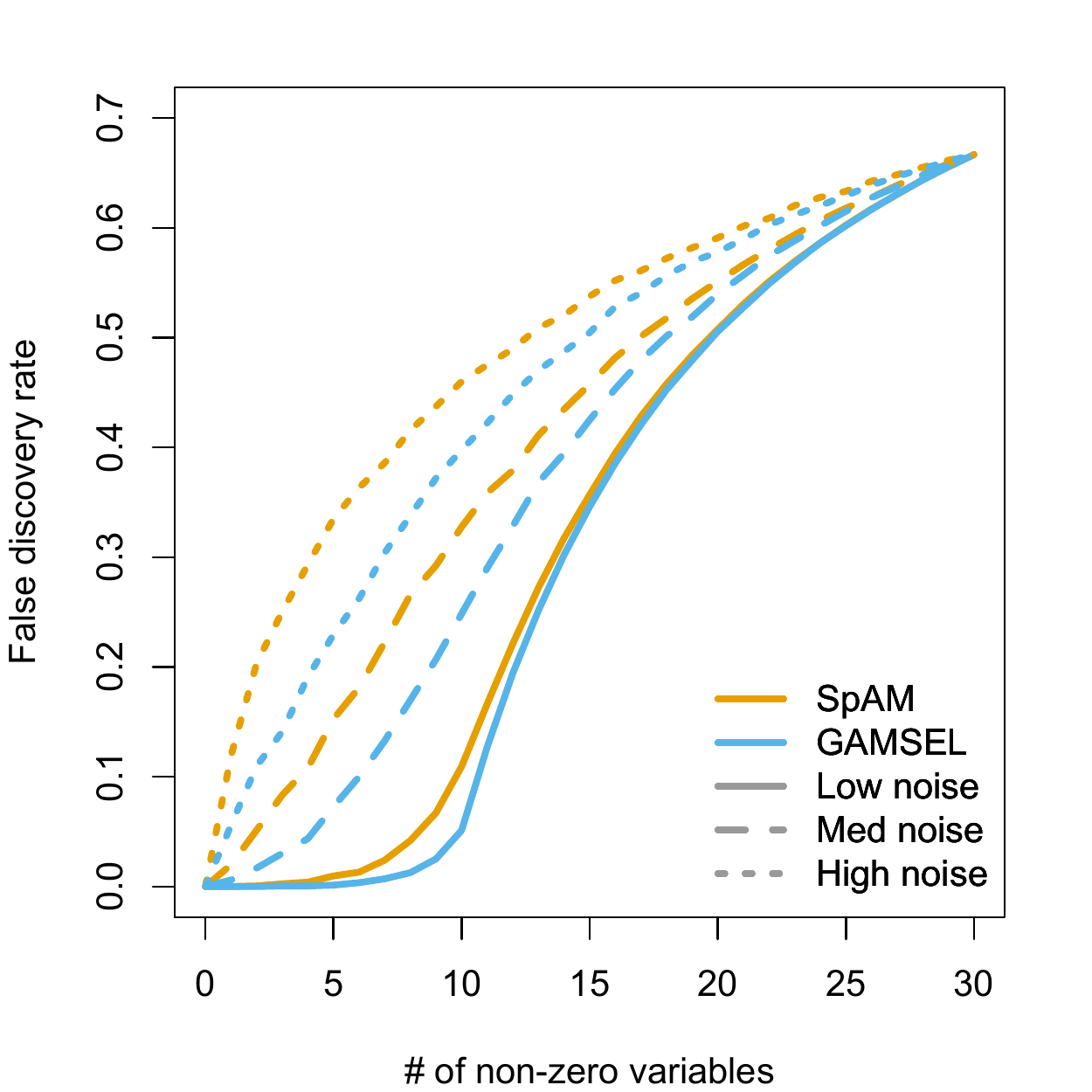}
  \caption{\#(lin, nonlin) = (10,0)}
  \end{subfigure}
  \begin{subfigure}[b]{0.32\textwidth}
  \includegraphics[width=\textwidth]{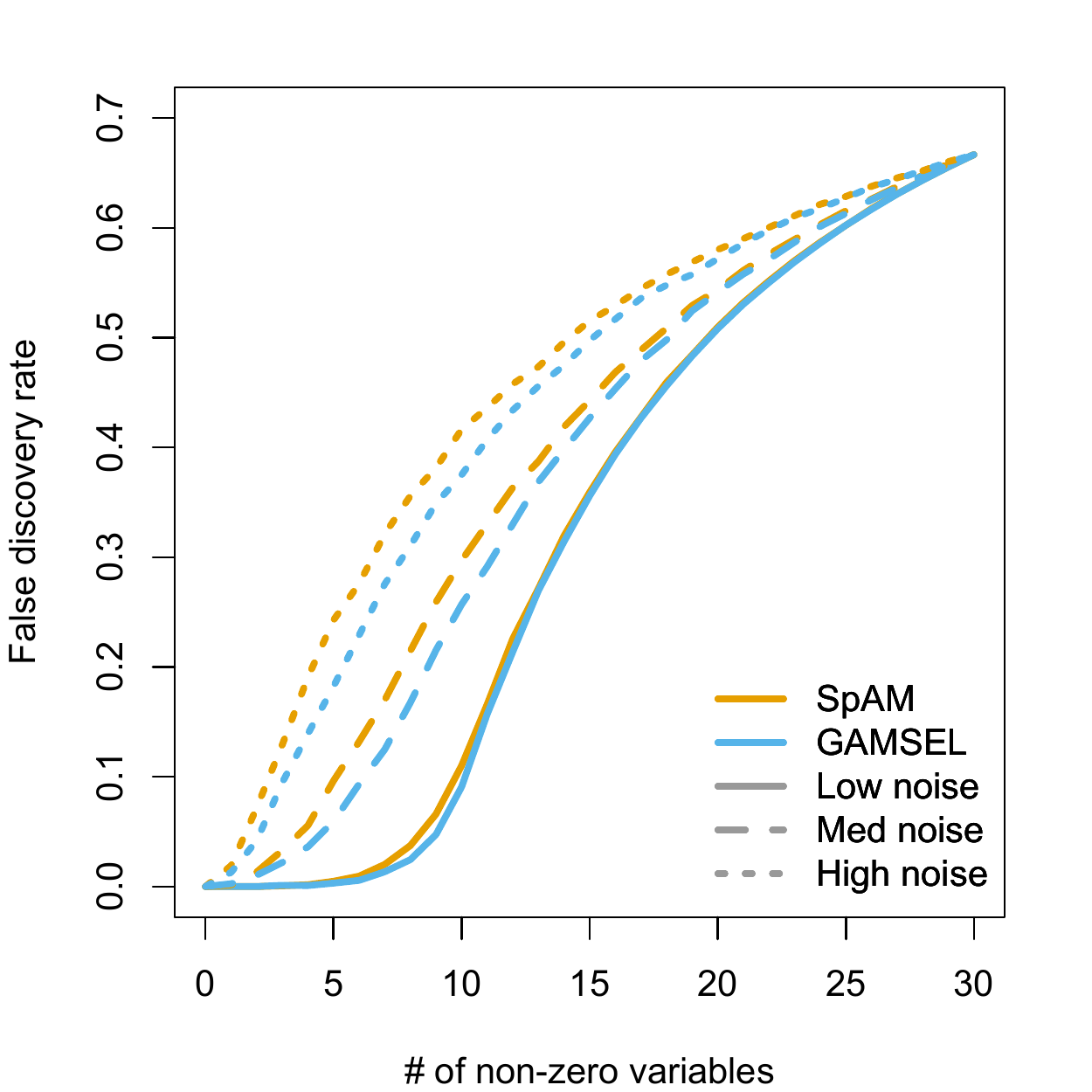}
  \caption{\#(lin, nonlin) = (6,4)}
  \end{subfigure}
  \begin{subfigure}[b]{0.32\textwidth}
  \includegraphics[width=\textwidth]{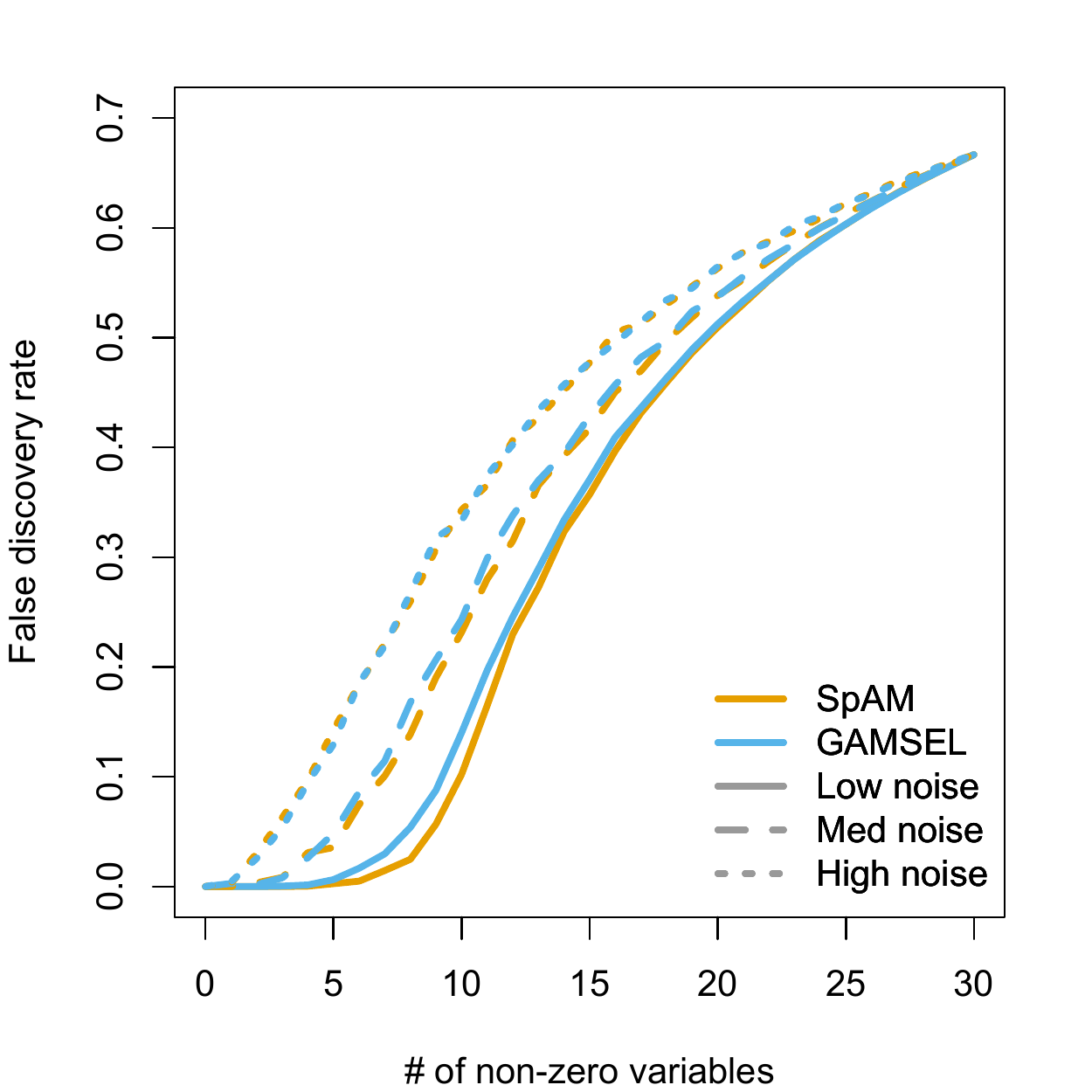}
  \caption{\#(lin, nonlin) = (0,10)}
  \end{subfigure}
  \caption{SpAM vs. GAMSEL ($\gamma = 0.5$) on simulated data example with $n=200$ observations and $p=30$ variables, of which $10$ are nonzero.  Three scenarios are considered: (a) all $10$ nonzero variables are linear; (b) 6 variables are linear, and 4 are nonlinear; (c) all 10 nonzero variables are nonlinear.  In cases (a) and (b) GAMSEL has a lower FDR than SpAM at all noise levels considered.  In case (c), where the ground truth is that all nonzero terms are nonlinear, SpAM slightly outperforms GAMSEL in the low noise regime.}
  \label{fig:fdrplot}
\end{figure}
We compared GAMSEL and SpAM based on the false discovery rate, defined as the fraction of variables estimated as non-zero that are truly $0$ in the underlying model.  To put the methods on equal footing, we compared their FDRs at equal model sizes.  Figure~\ref{fig:fdrplot} shows a summary of the results.  Six curves are shown in each subplot, one for each method across three different values of the noise variance.  In this comparison we find that GAMSEL performs considerably better in cases (a) and (b), where some of the nonzero variables are linear. The only case where SpAM is observed to outperform GAMSEL is in the low noise regime in case (c), where all nonzero effects are nonlinear.   Though we do not present the results here, we note that this trend was observed to persist at other values of noise variance and problem size (i.e., other choices of  $n$,$p$,\# linear, and \# nonlinear).

Next we revisit the spam dataset and use it to compare the prediction performance of GAMSEL and SpAM.  For this comparison we use the simulation setup described in the second part of section \ref{sec:spamdata}.  Starting with the full set of $4601$ messages, we randomly sample $300$ of them for use in training, and set the remaining $4301$ aside for testing.  The sampling is repeated $50$ times.

To put the methods on equal footing, we once again make the comparison in terms of the size of the selected model along the path, rather than the underlying value of the regularization parameter.  Since several penalty parameters can correspond to the same selected model, we record for each simulation instance the \emph{minimum} test set misclassification error at each size of selected model.  

\begin{figure}[!ht]
  \centering
    \includegraphics[width=\textwidth]{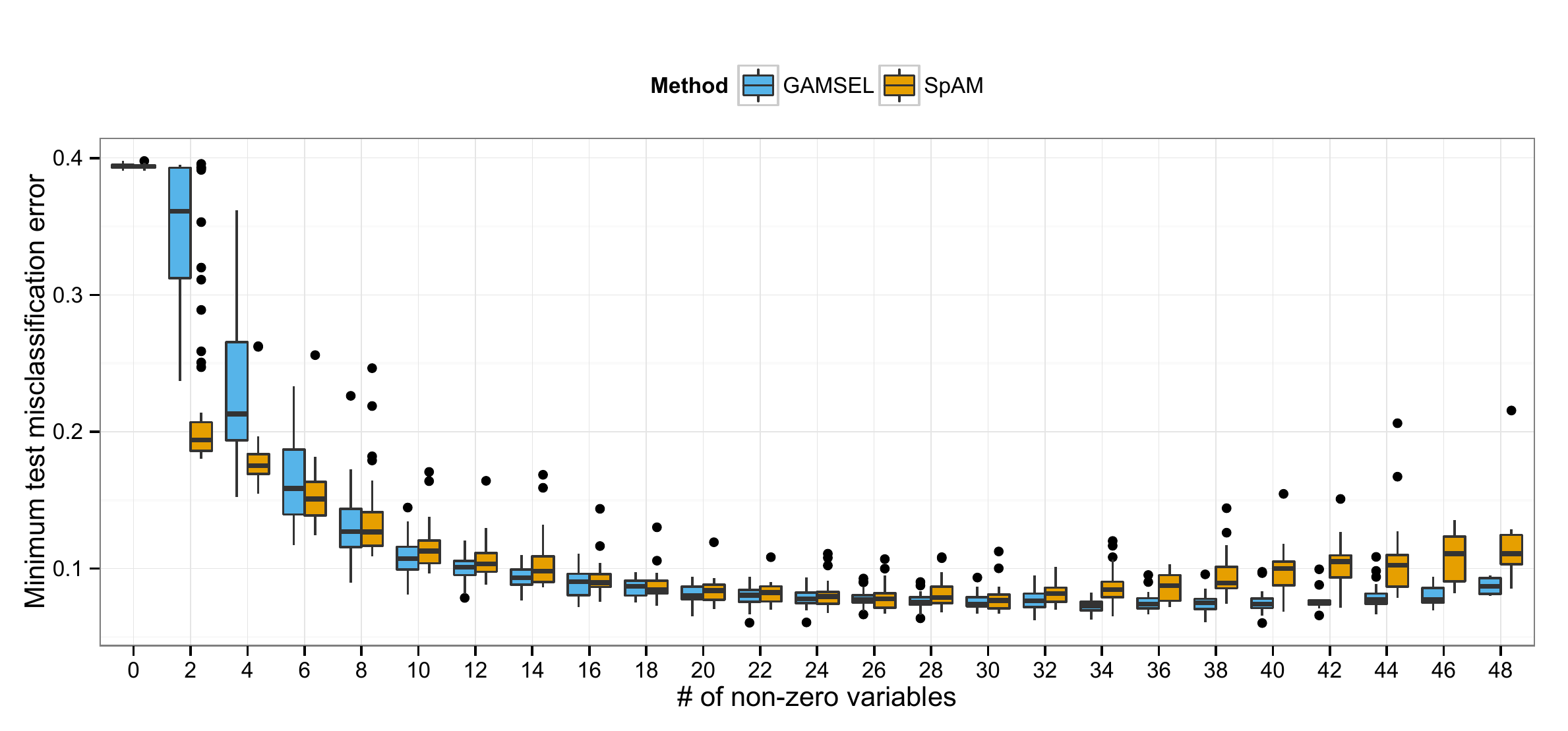}
    \caption{Test set misclassification error rates for GAMSEL and SpAM.  For display purposes, results are shown at only even values of selected model size.}
    \label{fig:spampredict}
\end{figure}

Figure~\ref{fig:spampredict} shows boxplots of the minimum test error at each (even) model size across the $50$ simulation iterations.  The main difference we observe comes at the extreme selected model sizes.  For small model sizes, SpAM tends to have a lower test error, while for large model sizes GAMSEL has lower test error.  The methods perform comparably in the middle range of model sizes.  A closer inspection would reveal that GAMSEL tends to achieve a lower minimum test error overall.   

\subsection{GAMSEL vs {\tt step.gam}}  

The {\tt step.gam} function in the {\tt gam} R package can be used to select between zero, linear and non-linear terms in generalized additive models via stepwise model selection.  For instance, instead of running GAMSEL with $5$ degrees of freedom, one could try running {\tt step.gam} initialized at the null model with scope {\tt \Twiddle 1 + x + s(x,5)} for each variable.  Since GAMSEL produces a fit with maximal degrees of freedom only at the end of the path when $\lambda = 0$, the analogy is not perfect.   Of course one could always add in formula terms of the form {\tt s(x,d)} for $1 < d < 5$, but doing so would increase computation time without greatly affecting model selection performance. 

The main disadvantage of {\tt step.gam} is the poor scalability of the procedure.  While GAMSEL scales to problems with thousands of observations and variables, {\tt step.gam} in general does not.  The search space quickly becomes too large for stepwise model selection to be computationally feasible.  In what follows we compare GAMSEL and step.gam both in terms of model selection performance and computation time.  

We begin by comparing the model selection performance of GAMSEL and {\tt step.gam} in the simulation setting described in \ref{sec:simstudy}.  We run {\tt step.gam} initialized at the null model with scope {\tt \Twiddle 1 + x + s(x,5)} for each variable, and compare it to GAMSEL with $10$ basis functions and $5$ degrees of freedom, taking $\gamma \in \{0.4, 0.5\}$.  The methods are compared on the basis of Zero vs. Nonzero, Linear and Nonlinear misclassification rates, as defined in \ref{sec:simstudy}.  To put both methods on equal footing the misclassification rates are compared at equal values of model size.  

\begin{figure}[!ht]
  \centering
  \begin{subfigure}[b]{0.4\linewidth}
    \includegraphics[width=\textwidth]{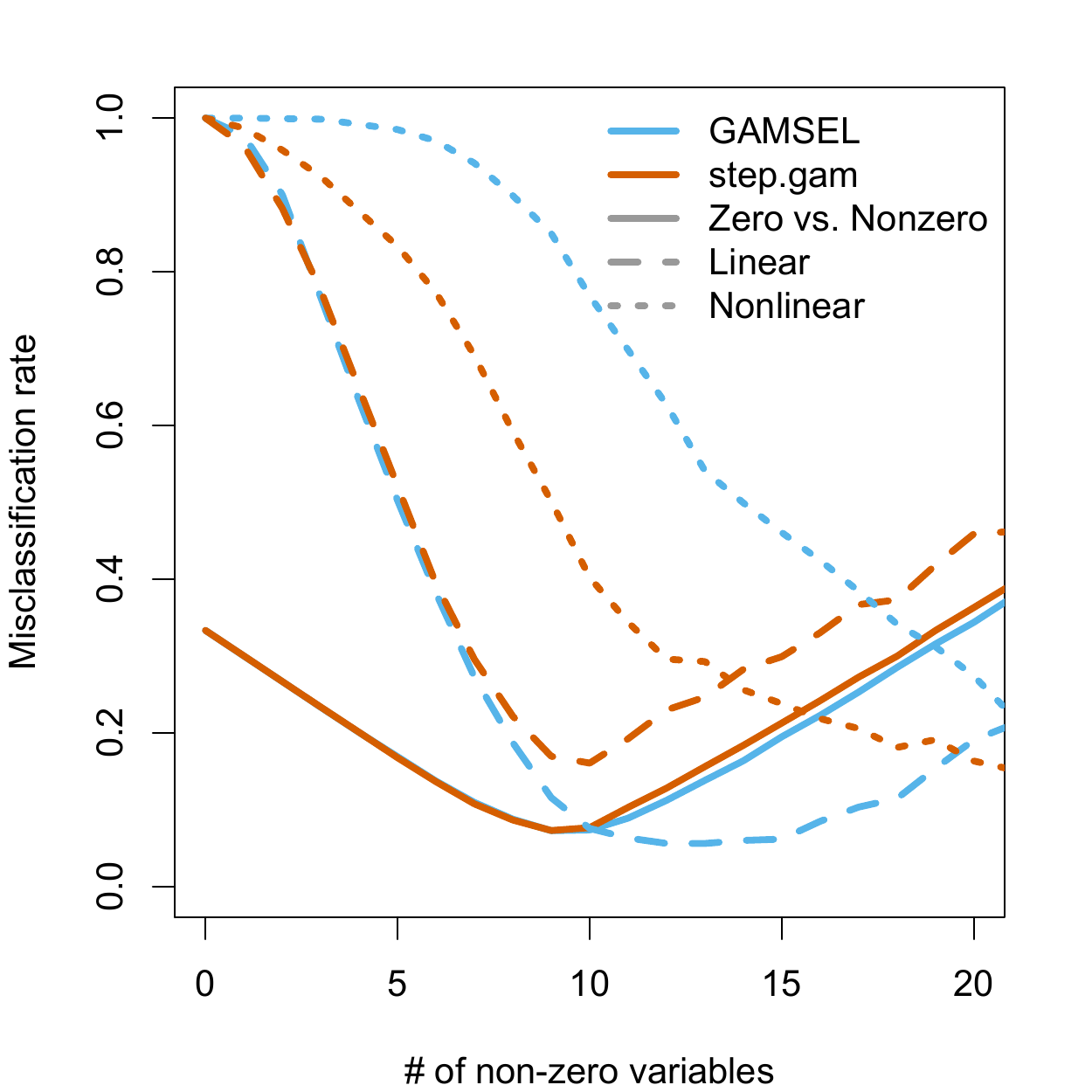}
    \caption{$\gamma = 0.4$}
  \end{subfigure}
  \begin{subfigure}[b]{0.4\linewidth}
    \includegraphics[width=\textwidth]{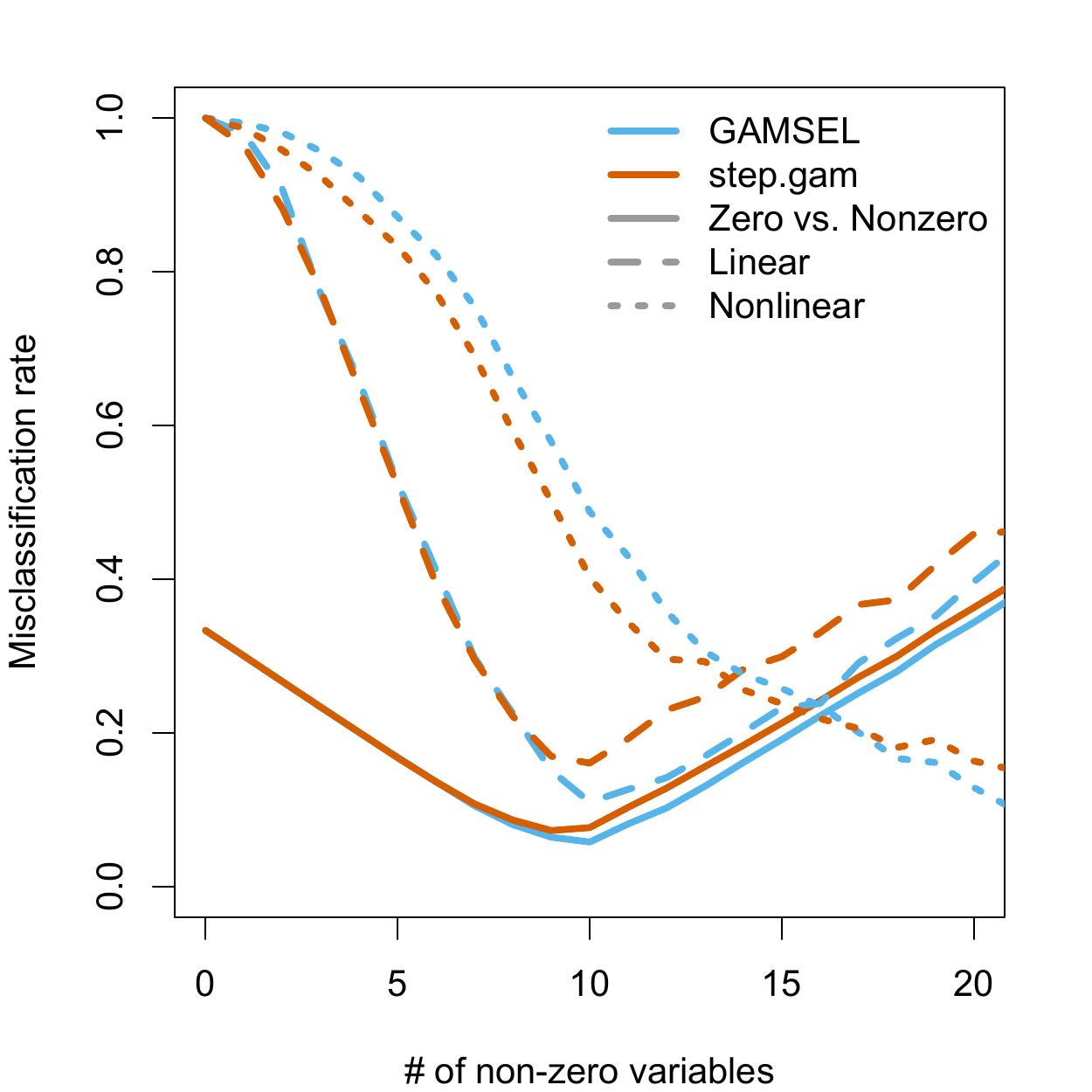}
    \caption{$\gamma = 0.5$}
  \end{subfigure}
  \caption{Classification performance comparison between {\tt step.gam} and GAMSEL.  Lower values are better.  Both methods have similar Zero vs. Nonzero misclassification rates, with GAMSEL performing slightly better when the selected model size exceeds $10$.  With $\gamma = 0.5$, the two methods also have very similar Linear and Nonlinear misclassification rates.  }
  \label{fig:gamselvsstep}
\end{figure}

Figure~\ref{fig:gamselvsstep} shows plots of misclassification rates for the two procedures.  GAMSEL and {\tt step.gam} perform comparably in terms of Zero vs. Nonzero misclassification rates.  In the case $\gamma = 0.5$, the two methods also have fairly similar Linear and Nonlinear misclassification rates across the range of model sizes. With $\gamma = 0.4$, GAMSEL has considerably lower Linear misclassification rates, but higher Nonlinear misclassification rates.  This trade-off between Linear/Nonlinear classification and interpretability may be a desirable one, provided that the cost in model fit is not too great. 

Next we report the results of a pair of timing experiments conducted to compare the computational performance of GAMSEL and {\tt step.gam}.  The first timing experiment reports the timings for the simulation described above.  For GAMSEL, we record the combined computation time expended on forming the bases $\{U_j\}$ and obtaining model fits for the full sequence of $\lambda$ values.  For {\tt step.gam} we record the computation time expended on doing $30$ steps of forward stepwise model selection. 

The left panel of Table \ref{tab:timing} shows a summary of the results.  Even on this small example {\tt step.gam} takes a fair bit of time to run.  In this comparison GAMSEL is over $30$ times faster.  

\begin{table}[h]
  \begin{tabular}{|c|c|}
    \hline
    \multicolumn{2}{|c|}{$n=200$, $p=30$} \\
    \hline
    GAMSEL & {\tt step.gam} \\ \hline
    $1.24 \pm 0.1$ & $41.7 \pm 0.6$ \\
    \hline
  \end{tabular}
  \hspace{5em}
  \begin{tabular}{|c|c|}
    \hline
    \multicolumn{2}{|c|}{$n=600$, $p=90$} \\
    \hline
    GAMSEL & {\tt step.gam} \\ \hline
    $9.1 \pm 0.7$  &606 $\pm 32$ \\
    \hline
  \end{tabular}
  \caption{Average computation time $\pm$ standard deviation (in seconds) for GAMSEL and {\tt step.gam} }
  \label{tab:timing}
\end{table}

For our second experiment we increase the problem size to $n=600$ observations and $p=90$ variables, of which $20$ are nonzero ($12$ linear, $8$ nonlinear).  We record computation times as before, still only running {\tt step.gam} for just $30$ steps of forward stepwise selection.  Results are reported in the right panel of Table \ref{tab:timing}.    It is clear that {\tt step.gam} does not scale well to larger problems.  In particular, while timing results shown are times for just the first $30$ steps, with $p = 90$ one would typically wish to consider a larger number of steps, which would further increase the computation time.  

We also note that roughly 67\% of the computation time of GAMSEL is expended on forming the bases $\{U_j\}$.  Once the $\{U_j\}$ are constructed, the average time expended computing the full solution path is $3.1$ seconds.  

\subsection{GAMSEL vs {\tt gam.selection \{mgcv\}}}  

Our final comparison is to the automatic term selection approach implemented in the {\tt mgcv} package in {\tt R} \citep{mgcvpkg}.
While the {\tt mgcv} package is most often used for automated smoothness selection rather than variable selection proper, the package also supports variable selection using an approach that augments the objective function with an additional penalty term \citep{marra2011practical}.  Unlike GAMSEL, this approach returns a single selected model rather than a sequence of models of increasing complexity.  To put the methods on equal footing, we compare the model selected using {\tt gam.selection} to the GAMSEL model selected via cross-validation using the 1-standard error rule.  We compare to each of the five {\tt method} options implemented in the {\tt mgcv} package.  

The simulation setup for our comparison is the same as in \ref{sec:simstudy}.  We run the {\tt gam} fitting function with the formula {\tt y $\sim$ 1 + s(X1, k=5, bs=`cr') + ... + s(X30, k=5, bs=`cr')} and flag {\tt select = TRUE}.  The choice of {\tt k = 5} is made to ensure that $n \cdot k < p$, which is a requirement of the {\tt mgcv} fit.  In interpreting the results, we deem that term in the {\tt mgcv} fit is zero if the edf is $< 0.1$, linear if the edf is in the range $[0.1, 1.3)$, and non-linear otherwise.  Our results are robust to the choice of cutoffs; just about any reasonable choice produces the same results.  Our findings are summarized in Table~\ref{tab:gamsel_vs_mgcv}. GAMSEL does a far better job in screening zero or linear terms than {\tt gam.selection}.

\begin{table}[ht]
\caption{
  Comparison of GAMSEL to {\tt mgcv} automated term selection approaches using the simulation setup described in \ref{sec:simstudy}.  Quantities are averages over 100 iterations of the simulation.  GAMSEL tends to select smaller models than any of the {\tt mgcv} methods, while having comparable recall and much higher precision on overall variable selection.  With the choice of $\gamma = 0.4$, GAMSEL applies a higher penalty to non-linear terms than to linear ones, which results in higher recall of linear terms compared to non-linaer terms.   GAMSEL considerably outperforms {\tt mgcv} on all measures except non-linear term recall.  }
\label{tab:gamsel_vs_mgcv}
\begin{center}
\begin{tabular}{|r|l|lllll|}
  \hline 
method & GAMSEL & GCV.Cp & ML & P-ML & P-REML & REML \\ 
\hline
  \# nonzero terms & 17 & 25 & 24 & 23 & 23 & 24 \\ 
  Zero-vs-nonzero misclassification rate & 0.25 & 0.51 & 0.47 & 0.44 & 0.44 & 0.47 \\ 
  Precision (nonzero terms) & 0.61 & 0.40 & 0.42 & 0.44 & 0.44 & 0.42 \\ 
  Recall (nonzero terms) & 0.97 & 0.99 & 0.99 & 0.99 & 0.99 & 0.99 \\ 
  Precision (linear terms) & 0.43 & 0.32 & 0.34 & 0.37 & 0.37 & 0.34 \\ 
  Recall (linear terms) & 0.86 & 0.52 & 0.74 & 0.77 & 0.76 & 0.74 \\ 
  Precision (nonlinear terms) & 0.69 & 0.24 & 0.32 & 0.33 & 0.33 & 0.32 \\ 
  Recall (nonlinear terms) & 0.61 & 0.90 & 0.84 & 0.82 & 0.82 & 0.83 \\ 
   \hline
\end{tabular}
  \end{center}
\end{table}

\section{\texttt{Gamsel} package in \texttt{R}}\label{sec:gamsel.pkg}
We have produced an \texttt{R} package \texttt{gamsel} that implements GAMSEL for squared-error loss and binomial log likelihood (logistic regression). The bulk of the fitting is done in C, which results in fast execution times. Similar in structure to the package \texttt{glmnet}, the \texttt{gamsel} function fits the entire regularization path for the GAMSEL objective (\ref{eq:objective}) over a grid of values of $\lambda$ (evenly spaced on the log scale; see~(\ref{eq:lambdamax})). 
There are \texttt{predict}, \texttt{plot} and \texttt{summary} methods for fitted GAMSEL objects, and \texttt{cv.gamsel} performs cross-validation for selecting the tuning parameter $\lambda$. The package is available from \texttt{http://cran.r-project.org/}.

\section{Conclusion}  \label{sec:conclusion}
In this paper we introduced GAMSEL, a penalized likelihood procedure for fitting sparse generalized additive models that scales to high-dimensional data.  Unlike competing methods, GAMSEL adaptively selects between zero, linear and non-linear fits for the component functions.  Our estimator therefore retains the interpretational advantages of linear fits when they provide a good description of the data, while still capturing any strong non-linearities that may be present.  

We investigate the model selection and prediction performance of our method through both simulated and real data examples.  In comparisons to SpAM, a competing method, our experiments indicate that GAMSEL can have notably better selection performance when some of the underlying effects are in truth linear.  Moreover, GAMSEL was observed to perform competitively even when the ground truth is that all effects are non-linear.  We also showed that our method performs comparably to forward stepwise selection as implemented in the {\tt gam} package, while also scaling well to larger problems.  

The block-wise coordinate descent algorithm for fitting the GAMSEL model in the case of linear and logistic regression, along with the plotting and cross-validation routines are implemented in the {\tt gamsel} package in R.

%


\bibliographystyle{imsart-nameyear}
\bibliography{gamsel}

\begin{thebibliography}{21}

\bibitem[\protect\citeauthoryear{Demmler and
  Reinsch}{1975}]{demmler75:_oscil_matric_with_splin_smoot}
\begin{barticle}[author]
\bauthor{\bsnm{Demmler},~\bfnm{A.}\binits{A.}} \AND
  \bauthor{\bsnm{Reinsch},~\bfnm{C.}\binits{C.}}
(\byear{1975}).
\btitle{Oscillation matrices with spline smoothing}.
\bjournal{Numerische Mathematik}
\bvolume{24}.
\end{barticle}
\endbibitem

\bibitem[\protect\citeauthoryear{Friedman, Hastie and
  Tibshirani}{2010}]{friedman08:_regul_paths_gener_linear_model_coord_descen}
\begin{barticle}[author]
\bauthor{\bsnm{Friedman},~\bfnm{Jerome}\binits{J.}},
  \bauthor{\bsnm{Hastie},~\bfnm{Trevor}\binits{T.}} \AND
  \bauthor{\bsnm{Tibshirani},~\bfnm{Robert}\binits{R.}}
(\byear{2010}).
\btitle{Regularization Paths for Generalized Linear Models via Coordinate
  Descent}.
\bjournal{Journal of Statistical Software}
\bvolume{33}
\bpages{1--22}.
\end{barticle}
\endbibitem

\bibitem[\protect\citeauthoryear{Green and
  Silverman}{1993}]{green1993nonparametric}
\begin{bbook}[author]
\bauthor{\bsnm{Green},~\bfnm{Peter~J}\binits{P.~J.}} \AND
  \bauthor{\bsnm{Silverman},~\bfnm{Bernard~W}\binits{B.~W.}}
(\byear{1993}).
\btitle{Nonparametric regression and generalized linear models: a roughness
  penalty approach}.
\bpublisher{CRC Press}.
\end{bbook}
\endbibitem

\bibitem[\protect\citeauthoryear{Hastie}{1995}]{Hastie95}
\begin{barticle}[author]
\bauthor{\bsnm{Hastie},~\bfnm{T.}\binits{T.}}
(\byear{1995}).
\btitle{Pseudosplines}.
\bjournal{Journal of the Royal Statistical Society, Series B}
\bvolume{58}
\bpages{379-396}.
\end{barticle}
\endbibitem

\bibitem[\protect\citeauthoryear{Hastie}{2015}]{gampkg}
\begin{bmanual}[author]
\bauthor{\bsnm{Hastie},~\bfnm{Trevor}\binits{T.}}
(\byear{2015}).
\btitle{gam: Generalized Additive Models}
\bnote{R package version 1.12}.
\end{bmanual}
\endbibitem

\bibitem[\protect\citeauthoryear{Hastie and
  Tibshirani}{1986}]{hastie1986generalized}
\begin{barticle}[author]
\bauthor{\bsnm{Hastie},~\bfnm{Trevor}\binits{T.}} \AND
  \bauthor{\bsnm{Tibshirani},~\bfnm{Robert}\binits{R.}}
(\byear{1986}).
\btitle{Generalized additive models}.
\bjournal{Statistical science}
\bpages{297--310}.
\end{barticle}
\endbibitem

\bibitem[\protect\citeauthoryear{Hastie and
  Tibshirani}{1990}]{hastie1990generalized}
\begin{bbook}[author]
\bauthor{\bsnm{Hastie},~\bfnm{Trevor~J}\binits{T.~J.}} \AND
  \bauthor{\bsnm{Tibshirani},~\bfnm{Robert~J}\binits{R.~J.}}
(\byear{1990}).
\btitle{Generalized additive models}
\bvolume{43}.
\bpublisher{CRC Press}.
\end{bbook}
\endbibitem

\bibitem[\protect\citeauthoryear{Hastie, Tibshirani and
  Wainwright}{2015}]{hastie15:_statis_learn_with_spars}
\begin{bbook}[author]
\bauthor{\bsnm{Hastie},~\bfnm{T.}\binits{T.}},
  \bauthor{\bsnm{Tibshirani},~\bfnm{R.}\binits{R.}} \AND
  \bauthor{\bsnm{Wainwright},~\bfnm{M.}\binits{M.}}
(\byear{2015}).
\btitle{Statistical Learning with Sparsity: the Lasso and Generalizations}.
\bpublisher{Chapman and Hall, CRC Press}.
\end{bbook}
\endbibitem

\bibitem[\protect\citeauthoryear{Hastie et~al.}{2009}]{hastie2009elements}
\begin{bbook}[author]
\bauthor{\bsnm{Hastie},~\bfnm{Trevor}\binits{T.}},
  \bauthor{\bsnm{Tibshirani},~\bfnm{Robert}\binits{R.}},
  \bauthor{\bsnm{Friedman},~\bfnm{Jerome}\binits{J.}},
  \bauthor{\bsnm{Hastie},~\bfnm{T}\binits{T.}},
  \bauthor{\bsnm{Friedman},~\bfnm{J}\binits{J.}} \AND
  \bauthor{\bsnm{Tibshirani},~\bfnm{R}\binits{R.}}
(\byear{2009}).
\btitle{The elements of statistical learning}
\bvolume{2}.
\bpublisher{Springer}.
\end{bbook}
\endbibitem

\bibitem[\protect\citeauthoryear{Jacob, Obozinski and
  Vert}{2009}]{jacob2009group}
\begin{binproceedings}[author]
\bauthor{\bsnm{Jacob},~\bfnm{Laurent}\binits{L.}},
  \bauthor{\bsnm{Obozinski},~\bfnm{Guillaume}\binits{G.}} \AND
  \bauthor{\bsnm{Vert},~\bfnm{Jean-Philippe}\binits{J.-P.}}
(\byear{2009}).
\btitle{Group lasso with overlap and graph lasso}.
In \bbooktitle{Proceedings of the 26th Annual International Conference on
  Machine Learning}
\bpages{433--440}.
\bpublisher{ACM}.
\end{binproceedings}
\endbibitem

\bibitem[\protect\citeauthoryear{Lin et~al.}{2006}]{lin2006component}
\begin{barticle}[author]
\bauthor{\bsnm{Lin},~\bfnm{Yi}\binits{Y.}},
  \bauthor{\bsnm{Zhang},~\bfnm{Hao~Helen}\binits{H.~H.}} \betal{et~al.}
(\byear{2006}).
\btitle{Component selection and smoothing in multivariate nonparametric
  regression}.
\bjournal{The Annals of Statistics}
\bvolume{34}
\bpages{2272--2297}.
\end{barticle}
\endbibitem

\bibitem[\protect\citeauthoryear{Lou et~al.}{2014}]{lou2014sparse}
\begin{barticle}[author]
\bauthor{\bsnm{Lou},~\bfnm{Yin}\binits{Y.}},
  \bauthor{\bsnm{Bien},~\bfnm{Jacob}\binits{J.}},
  \bauthor{\bsnm{Caruana},~\bfnm{Rich}\binits{R.}} \AND
  \bauthor{\bsnm{Gehrke},~\bfnm{Johannes}\binits{J.}}
(\byear{2014}).
\btitle{Sparse partially linear additive models}.
\bjournal{arXiv preprint arXiv:1407.4729}.
\end{barticle}
\endbibitem

\bibitem[\protect\citeauthoryear{Marra and Wood}{2011}]{marra2011practical}
\begin{barticle}[author]
\bauthor{\bsnm{Marra},~\bfnm{Giampiero}\binits{G.}} \AND
  \bauthor{\bsnm{Wood},~\bfnm{Simon~N}\binits{S.~N.}}
(\byear{2011}).
\btitle{Practical variable selection for generalized additive models}.
\bjournal{Computational Statistics \& Data Analysis}
\bvolume{55}
\bpages{2372--2387}.
\end{barticle}
\endbibitem

\bibitem[\protect\citeauthoryear{Marx and
  Eilers}{1998}]{marx98:_direc_gener_addit_model_with_penal_likel}
\begin{barticle}[author]
\bauthor{\bsnm{Marx},~\bfnm{B.}\binits{B.}} \AND
  \bauthor{\bsnm{Eilers},~\bfnm{P.}\binits{P.}}
(\byear{1998}).
\btitle{Direct generalized additive modeling with penalized likelihood}.
\bjournal{Computational Statistics and Data Analysis}
\bvolume{28}
\bpages{193-209}.
\end{barticle}
\endbibitem

\bibitem[\protect\citeauthoryear{Meier et~al.}{2009}]{meier2009high}
\begin{barticle}[author]
\bauthor{\bsnm{Meier},~\bfnm{Lukas}\binits{L.}}, \bauthor{\bparticle{Van~de}
  \bsnm{Geer},~\bfnm{Sara}\binits{S.}},
  \bauthor{\bsnm{B{\"u}hlmann},~\bfnm{Peter}\binits{P.}} \betal{et~al.}
(\byear{2009}).
\btitle{High-dimensional additive modeling}.
\bjournal{The Annals of Statistics}
\bvolume{37}
\bpages{3779--3821}.
\end{barticle}
\endbibitem

\bibitem[\protect\citeauthoryear{Ravikumar et~al.}{2007}]{ravikumar2007spam}
\begin{binproceedings}[author]
\bauthor{\bsnm{Ravikumar},~\bfnm{Pradeep~D}\binits{P.~D.}},
  \bauthor{\bsnm{Liu},~\bfnm{Han}\binits{H.}},
  \bauthor{\bsnm{Lafferty},~\bfnm{John~D}\binits{J.~D.}} \AND
  \bauthor{\bsnm{Wasserman},~\bfnm{Larry~A}\binits{L.~A.}}
(\byear{2007}).
\btitle{SpAM: Sparse Additive Models.}
In \bbooktitle{NIPS}.
\end{binproceedings}
\endbibitem

\bibitem[\protect\citeauthoryear{Tibshirani}{1996}]{tibshirani1996regression}
\begin{barticle}[author]
\bauthor{\bsnm{Tibshirani},~\bfnm{Robert}\binits{R.}}
(\byear{1996}).
\btitle{Regression shrinkage and selection via the lasso}.
\bjournal{Journal of the Royal Statistical Society. Series B (Methodological)}
\bpages{267--288}.
\end{barticle}
\endbibitem

\bibitem[\protect\citeauthoryear{Tibshirani
  et~al.}{2012}]{tibshirani2012strong}
\begin{barticle}[author]
\bauthor{\bsnm{Tibshirani},~\bfnm{Robert}\binits{R.}},
  \bauthor{\bsnm{Bien},~\bfnm{Jacob}\binits{J.}},
  \bauthor{\bsnm{Friedman},~\bfnm{Jerome}\binits{J.}},
  \bauthor{\bsnm{Hastie},~\bfnm{Trevor}\binits{T.}},
  \bauthor{\bsnm{Simon},~\bfnm{Noah}\binits{N.}},
  \bauthor{\bsnm{Taylor},~\bfnm{Jonathan}\binits{J.}} \AND
  \bauthor{\bsnm{Tibshirani},~\bfnm{Ryan~J}\binits{R.~J.}}
(\byear{2012}).
\btitle{Strong rules for discarding predictors in lasso-type problems}.
\bjournal{Journal of the Royal Statistical Society: Series B (Statistical
  Methodology)}
\bvolume{74}
\bpages{245--266}.
\end{barticle}
\endbibitem

\bibitem[\protect\citeauthoryear{Wood}{2011}]{wood2011fast}
\begin{barticle}[author]
\bauthor{\bsnm{Wood},~\bfnm{Simon~N}\binits{S.~N.}}
(\byear{2011}).
\btitle{Fast stable restricted maximum likelihood and marginal likelihood
  estimation of semiparametric generalized linear models}.
\bjournal{Journal of the Royal Statistical Society: Series B (Statistical
  Methodology)}
\bvolume{73}
\bpages{3--36}.
\end{barticle}
\endbibitem

\bibitem[\protect\citeauthoryear{Wood}{2015}]{mgcvpkg}
\begin{bmanual}[author]
\bauthor{\bsnm{Wood},~\bfnm{Simon}\binits{S.}}
(\byear{2015}).
\btitle{Mixed GAM Computation Vehicle with GCV/AIC/REML Smoothness Estimation}
\bnote{R package version 1.8-6}.
\end{bmanual}
\endbibitem

\bibitem[\protect\citeauthoryear{Yuan and Lin}{2006}]{yuan2006model}
\begin{barticle}[author]
\bauthor{\bsnm{Yuan},~\bfnm{Ming}\binits{M.}} \AND
  \bauthor{\bsnm{Lin},~\bfnm{Yi}\binits{Y.}}
(\byear{2006}).
\btitle{Model selection and estimation in regression with grouped variables}.
\bjournal{Journal of the Royal Statistical Society: Series B (Statistical
  Methodology)}
\bvolume{68}
\bpages{49--67}.
\end{barticle}
\endbibitem

\end{thebibliography}

\end{document}